\documentclass[10pt,twocolumn,letterpaper]{article}

\usepackage[applications]{wacv}      %

\usepackage{graphicx}
\usepackage{amsmath}
\usepackage{amssymb}
\usepackage{multirow}
\usepackage{booktabs}
\usepackage{xspace}
\usepackage{colortbl} %

\usepackage[table]{xcolor}
\usepackage{enumitem}

\usepackage[pagebackref,breaklinks,colorlinks]{hyperref}

\usepackage{appendix}
\usepackage[capitalize]{cleveref}
\crefname{section}{Sec.}{Secs.}
\Crefname{section}{Section}{Sections}
\Crefname{table}{Table}{Tables}
\crefname{table}{Tab.}{Tabs.}

\def\wacvPaperID{379} %
\def\confName{WACV}
\def\confYear{2025}

\newcommand{\PS}{\textsc{ProSkill}\xspace}

\begin{document}

\title{\PS: Segment-Level Skill Assessment in Procedural Videos}

\author{Michele Mazzamuto\thanks{Equal contribution.}\\
University of Catania, Italy\\
\and
Daniele Di Mauro\footnotemark[1]\\
Next Vision s.r.l., Italy\\
\and
Gianpiero Francesca\thanks{Co-Principal Investigator role.}\\
Toyota Motor Europe, Belgium\\
\and
Giovanni Maria Farinella\footnotemark[2]\\
University of Catania, Italy\\
\and
Antonino Furnari\footnotemark[2]\\
University of Catania, Italy\\
}
\maketitle

\begin{abstract}
\noindent    Skill assessment in procedural videos is crucial for the objective evaluation of human performance in settings such as manufacturing and procedural daily tasks. Current research on skill assessment has predominantly focused on sports and lacks large-scale datasets for complex procedural activities.
Existing studies typically involve only a limited number of actions, focus on either pairwise assessments (e.g., A is better than B) or on binary labels (e.g., good execution vs needs improvement). 
In response to these shortcomings, we introduce \PS, the first benchmark dataset for action-level skill assessment in procedural tasks. \PS provides absolute skill assessment annotations, along with pairwise ones. This is enabled by a novel and scalable annotation protocol that allows for the creation of an absolute skill assessment ranking starting from pairwise assessments. This protocol leverages a Swiss Tournament scheme for efficient pairwise comparisons, which are then aggregated into consistent, continuous global scores using an ELO-based rating system. 
We use our dataset to benchmark the main state-of-the-art skill assessment algorithms, including both ranking-based and pairwise paradigms.
The suboptimal results achieved by the current state-of-the-art highlight the challenges and thus the value of \PS in the context of skill assessment for procedural videos. All data and code are available at \url{https://fpv-iplab.github.io/ProSkill/}.
\end{abstract}

\section{Introduction}
\label{sec:intro}
\begin{figure}[ht]
    \centering
    \includegraphics[width=0.47\textwidth]{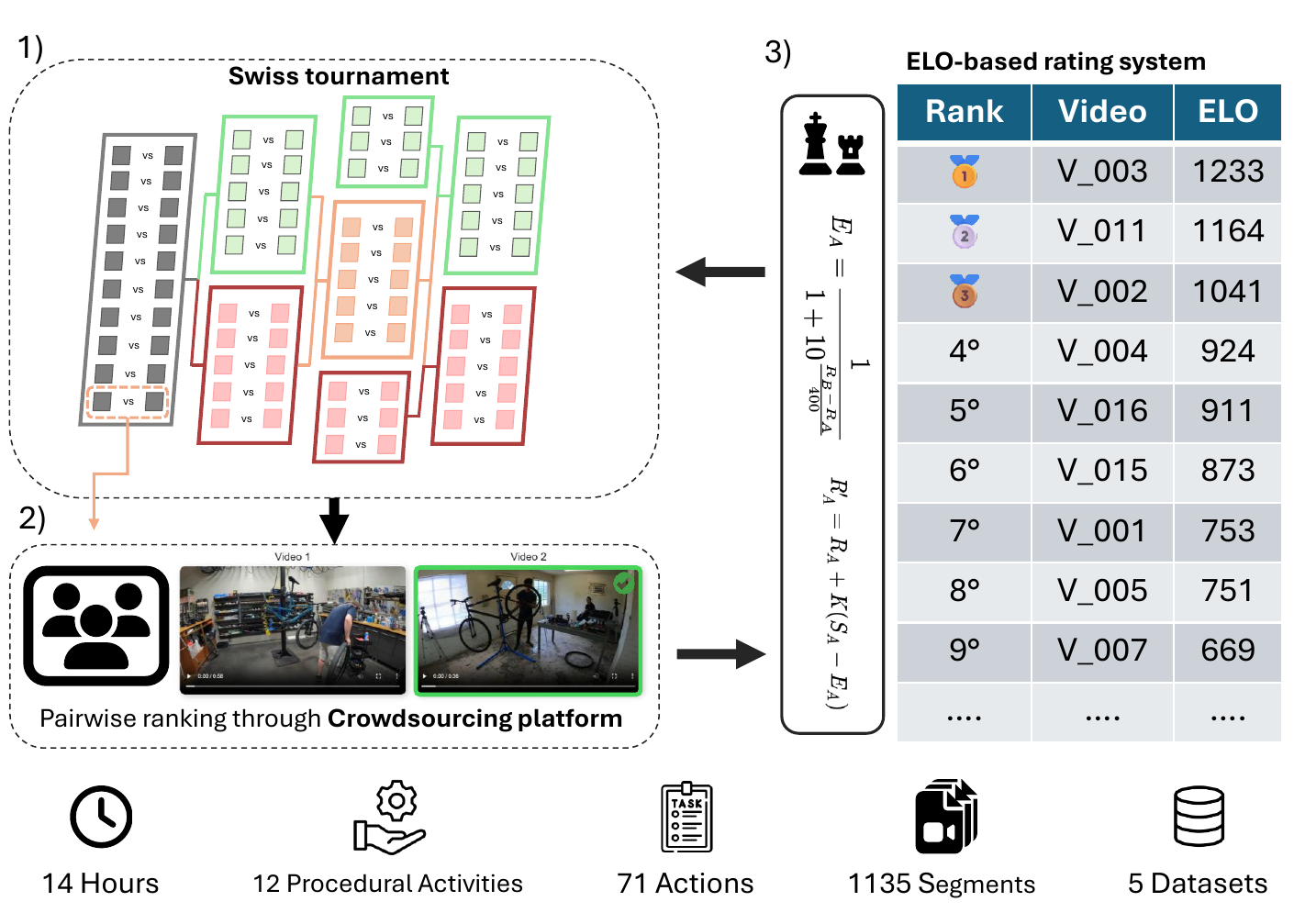}
    \caption{
        The proposed annotation protocol (top) and key features of the resulting \PS dataset (bottom). 
        \textbf{Stage 1} action video pairs are selected for labeling following a Round of a Swiss Tournament scheme;
        \textbf{Stage 2} selected pairs are labeled with a crowdsourcing platform asking users to perform pairwise ranking; 
        \textbf{Stage 3} the pairwise outcomes are aggregated using an ELO~\cite{elo1978rating} scheme to compute a global leaderboard.
        Stages 1-3 are iterated for a given number of rounds to achieve stable absolute ratings.
    }
    \label{fig:teaser}
\end{figure}
\noindent Skill assessment is a fundamental task in human activity understanding, as it enables the objective evaluation of how well a person performs a given task. This capability is particularly valuable in procedural settings such as manufacturing and assembly, where the assessment of employee skills can directly impact efficiency, safety, and quality. In addition, wearable devices offer a promising avenue for skill assessment, enabling continuous, real-time monitoring of performance for early detection of errors or suboptimal behavior~\cite{Mazzamuto_2025_CVPR,seminara2025differentiabletaskgraphlearning,sener2022assembly101,flaborea2024pregoonlinemistakedetection}. Previous research on human skill assessment has been predominantly focused on sports~\cite{xu2019learning,mtlaqa,xu2022finediving,parmar2022domain,data3010002}, where datasets and evaluation protocols are more mature and readily available. Some efforts have extended skill assessment to other contexts, including surgical procedures~\cite{doughty2019pros,liu2021unifiedsurgicalskillassessment,8354185} and egocentric activities~\cite{huang2024egoexolearn}. Unfortunately, these datasets are generally limited in scale, task diversity, and procedural structure. A recent step forward is represented by Ego-Exo4D~\cite{grauman_ego-exo4d_2024}, which captures goal-driven activities from both egocentric and exocentric perspectives, providing segment-level binary skill labels for a subset of videos. 
Despite these efforts, a large-scale, annotated dataset offering skill-level annotations across diverse real-world tasks is currently missing. 

To address this limitation, we introduce \PS (Figure~\ref{fig:teaser}) the first dataset specifically designed for procedural skill assessment, with the aim of supporting the development and evaluation of models that estimate human expertise in  structured tasks. 
Previous skill assessment datasets have typically adopted either absolute scores~\cite{10.1007/978-3-319-10599-4_36,gao2014jhu,parmar2017learning,parmar2019action,mtlaqa,xu2019learning,sener2022assembly101} or pairwise comparisons~\cite{doughty2018s, doughty2019pros,huang2024egoexolearn} to asses skill. Absolute scoring allows direct skill estimation and supports regression-based evaluation, but is often ambiguous and inconsistent outside of sports and other constrained domains where trained judges can apply objective criteria (e.g., body positions). Pairwise comparisons, on the other hand, are more intuitive and reliable for annotators, as they reduce ambiguity by asking to evaluate which of two performances is better. However, they do not inherently provide globally interpretable skill scores, which are essential for actionable feedback (e.g., “above average” or “needs improvement”).
\PS bridges this gap by combining pairwise comparisons into globally consistent ranking, allowing for multiple evaluation paradigms (ranking, regression, and classification), within a dataset. 

To achieve this, we propose a novel annotation protocol consisting of three stages based on a Swiss Tournament scheme, crowdsourcing annotation and ELO rating system~\cite{elo1978rating}. Swiss Tournament and ELO are long-established practices in competitive domains such as chess. This approach avoids the cost of exhaustive pairwise comparisons 
while preserving robustness. %
\PS consists of recordings of diverse tasks performed in unconstrained, real-world conditions. Such recordings are gathered from publicly available video datasets that have been widely used for procedural activity understanding. Specifically, \PS includes curated annotations on sequences from \emph{EgoExo4D}~\cite{grauman_ego-exo4d_2024}, \emph{Meccano}~\cite{ragusa2021meccano}, \emph{EpicTent}~\cite{jang2019epic}, \emph{Ikea ASM}~\cite{ben-shabat_ikea_2023}, and \emph{Assembly101}~\cite{sener2022assembly101}.These datasets cover diverse tasks, environments, and viewpoints, making \PS a comprehensive benchmark for skill assessment across multiple domains. Using \PS, we evaluate several state-of-the-art algorithms, including both global- and pairwise-ranking models. Results indicate significant room for improvement, confirming that procedural skill assessment remains a challenging open problem.

In summary, the main contributions of this work are: 
1) A novel annotation protocol based on a Swiss Tournament, combining crowd-sourced pairwise annotations and ELO rating for scalable collection of absolute skill scores; 
2) \PS, the first dataset of procedural activities spanning multiple domains and environments, with both absolute and relative skill annotations; 
3) A benchmark of state-of-the-art methods for pairwise and global ranking on diverse real-world tasks. 
All labels, code for the annotation protocol, and experimental pipelines will be publicly released.

\begin{table*}[t]
\centering
\resizebox{0.9\textwidth}{!}{%
\begin{tabular}{lcccccccc}
\toprule
\multirow[b]{2}{*}{\textbf{Name}} & \multirow[b]{2}{*}{\textbf{Year}} & \multirow[b]{2}{*}{\textbf{Domain}} & \multicolumn{3}{c}{\textbf{Skill Annotation}} & \multirow[b]{2}{*}{\textbf{Samples}} & \multirow[b]{2}{*}{\textbf{Hours}} & \multirow[b]{2}{*}{\textbf{Actions}} \\
\cmidrule(lr){4-6}
& & & \textbf{Type} & \textbf{Source} & \textbf{Granularity} & & & \\
\midrule
\multicolumn{9}{l}{\textit{Datasets with Absolute Score Annotations}} \\
\midrule
MIT-Dive~\cite{10.1007/978-3-319-10599-4_36}    & 2014 & Sport         & Absolute Score   & Judges          & Video     & 159    & 0.12     & 1$\ast$  \\
MIT-Skate~\cite{10.1007/978-3-319-10599-4_36}   & 2014 & Sport         & Absolute Score   & Judges          & Video     & 150    & 7.29     & 1        \\
JIGSAWS~\cite{gao2014jhu}                       & 2014 & Surgery       & Absolute Score   & Experts         & Video     & 103    & $\sim$10 & 3        \\
UNLV-Dive~\cite{parmar2017learning}             & 2017 & Sport         & Absolute Score   & Judges          & Video     & 717    & 7.10     & 1$\ast$  \\
UNLV-Vault~\cite{parmar2017learning}            & 2017 & Sport         & Absolute Score   & Judges          & Video     & 176    & 0.12     & 1        \\
AQA-7~\cite{parmar2019action}                   & 2019 & Sport         & Absolute Score   & Judges          & Video     & 1106   & $\sim$2  & 7        \\
MTL-AQA~\cite{mtlaqa}                           & 2019 & Sport         & Absolute Score   & Judges          & Video     & 1412   & 1.6      & 2$\ast$  \\
Fis-V~\cite{xu2019learning}                     & 2019 & Sport         & Absolute Score   & Judges          & Video     & 500    & 22       & 1        \\
Assembly101~\cite{sener2022assembly101}         & 2022 & Assembly      & Absolute Score   & Self-reported   & Subject   & 53     & 513      & 202      \\
\midrule
\multicolumn{9}{l}{\textit{Datasets with Other Annotation Types (Uncertainty level, Difficulty, Categorical, Binary, etc.)}} \\
\midrule
EpicTent~\cite{jang2019epic}                    & 2019 & Assembly      & Uncertainty        & Self-reported   & Segment     & 34090     & 5.40     & 38       \\
Piano-Skills~\cite{parmar2021piano}             & 2021 & Music         & Grade, Difficulty  & Teachers        & Video     & 992    & $\sim$3  & 9$\star$ \\
HoloAssist~\cite{wang_holoassist_2023}          & 2023 & Manipulation  & Self-Reported Skill& Self-reported   & Subject   & 222    & 169      & 414      \\
\multirow{2}{*}{Ego-Exo4D~\cite{grauman_ego-exo4d_2024}} & \multirow{2}{*}{2024} & \multirow{2}{*}{Assembly + Daily} & Categorical Labels & Surveys & Subject & 740 & \multirow{2}{*}{1286} & \multirow{2}{*}{NA} \\
& & & Binary Labels & Experts & Segment & 2539 & & \\
\midrule
\multicolumn{9}{l}{\textit{Datasets with Pairwise Ranking Annotations}} \\
\midrule
EPIC-Skill~\cite{doughty2018s}                  & 2018 & Surgery + Daily & Pairwise Ranking & Crowdsourcing   & Video     & 216    & 5.20     & 4        \\
BEST~\cite{doughty2019pros}                     & 2019 & Daily         & Pairwise Ranking & Crowdsourcing   & Video     & 500    & 26       & 5        \\
EgoExoLearn~\cite{huang2024egoexolearn}         & 2024 & Cooking       & Pairwise Ranking & Crowdsourcing   & Segment   & 3304   & $\sim$9  & 4        \\
\midrule[\heavyrulewidth]
\rowcolor{gray!15}
\textbf{\PS} & \textbf{2025} & \textbf{Procedural} & \bfseries Absolute + Pairwise & \bfseries Crowdsourcing & \bfseries Segment & \bfseries 1135 & \bfseries 14 & \bfseries 71 \\
\bottomrule
\end{tabular}
}
\caption{Datasets with skill annotations. \textbf{\PS} provides fine-grained \textbf{segment-level} annotations on procedural videos with both \textbf{absolute and pairwise} skill ratings, covering \textbf{$14$ hours} of video and \textbf{$71$ diverse actions}. 
$\ast$: Dive height. $\star$: Song difficulty levels.}
\label{tab:skill_datasets_filtered}
\end{table*}

\section{Related Works}

\noindent
\textbf{Datasets for Skill Assessment}
Skill assessment from video has been explored across a variety of domains, including sports, medical training, rehabilitation, music, and procedural tasks. The resulting datasets differ significantly in their annotation strategies, ranging from absolute scores to pairwise comparisons, as well as in activity structure and data modality. This diversity reflects a fragmented landscape, where progress in one area often does not transfer easily to others.
Initial efforts concentrated on competitive sports, where performance is inherently linked to objective scoring. Datasets such as MIT-Dive~\cite{10.1007/978-3-319-10599-4_36}, MIT-Skate~\cite{10.1007/978-3-319-10599-4_36}, UNLV-Dive~\cite{parmar2017learning}, UNLV-Vault~\cite{parmar2017learning}, AQA-7~\cite{parmar2019action}, and MTL-AQA~\cite{mtlaqa} associate skill with scores assigned during official competitions. More recent datasets such as FineDiving~\cite{xu2022finediving}, LOGO~\cite{LOGO}, and FineFS~\cite{JI2023FineFS} improve annotation consistency by leveraging expert consensus, multi-view video, and more diverse subjects and actions.
Beyond sports, structured domains such as surgery and rehabilitation have received attention due to their reliance on expert-defined protocols and high-stakes outcomes. JIGSAWS~\cite{gao2014jhu} and KIMORE~\cite{JI2023FineFS}, for example, provide fine-grained demonstrations of surgical or rehabilitative tasks annotated using clinical frameworks like OSATS~\cite{OSATS10.1046/j.1365-2168.1997.02502.x}. These datasets often include multimodal data (e.g., RGB-D, motion capture), supporting detailed analysis of motor control and execution.
Moving further into everyday and creative domains, datasets such as EPIC-Skills~\cite{doughty2018s}, BEST~\cite{doughty2019pros}, and Piano-Skills~\cite{parmar2021piano} sidestep the subjectivity and cost of absolute scoring by adopting pairwise comparisons. This approach simplifies annotation and has been shown to scale effectively while maintaining robustness in skill ranking.

More recently, focus has shifted toward procedural and assembly tasks, where activities are long, hierarchical, and goal-directed. These tasks demand a deeper understanding of temporal structure, decision points, and success criteria. Assembly101~\cite{sener2022assembly101} provides egocentric recordings with subject-level self-assessments. EPIC-Tent~\cite{jang2019epic} captures egocentric tent assembly performances annotated with self-rated uncertainty scores. EgoExoLearn~\cite{huang2024egoexolearn} adopts pairwise comparisons over video segments to capture relative proficiency. The Struggle Determination dataset~\cite{feng2024strugglingdatasetbaselinesstruggle} takes a different perspective by labeling segments where users appear to struggle, surfacing cues like hesitation or error that are indicative of lower skill.
While these datasets mark progress toward real-world skill assessment in procedural videos, they have several limitations: small scale (few subjects or short videos), limited scenarios (constrained environments or narrow tasks), coarse labels (binary or clip-level instead of continuous or subject-level), low action variability, and lack of scalable annotation protocols (relying on self-reports or expert judgments). These factors limit generalization across tasks and hinder large-scale extension.\\
\noindent
Differently from prior efforts, \PS focuses on procedural activities annotated at the segment level, enabling more fine-grained assessment of user performance. 
It provides both pairwise skill comparisons and globally consistent absolute scores, covering multiple real-world domains.
As reported in Table \ref{tab:skill_datasets_filtered}, 
\PS is diverse, including $1135$ segments across $71$ actions and $14$ different hours of video, exceeding similar features of previous datasets related to sports, surgery, or music domains. 
While some benchmarks provide larger numbers of hours (e.g., Ego-Exo4D, Assembly101 and HoloAssist), or segments (e.g., EPIC-Tent and Ego-Exo4D), they either tackle the coarser-grained subject level, provide only binary labels, cover a very small number of actions, or report subjective labels (e.g., self-rated uncertainty) that limit their applicability.
All in all, \PS offers a unique combination of scale, diversity, and annotation depth, setting a new standard for the fine-grained analysis of procedural skills.

\noindent
\textbf{Global Ranking Methods}
\noindent Early Action Quality Assessment (AQA) approaches, such as Parmar et al.~\cite{parmar2017learningscoreolympicevents}, leveraged C3D features with SVR or LSTM regressors, aiming to balance data efficiency with temporal modeling. To mitigate label ambiguity, USDL~\cite{usdl} proposed modeling scores as probability distributions rather than point estimates, extended by MUSDL to incorporate multi-path supervision. TSA-Net~\cite{TSA-Net} improves contextual understanding through Tube Self-Attention focused on foreground motion, achieving efficient and competitive performance. DAE-AQA~\cite{10.1007/dae-aqa} adopts a variational autoencoder to learn score distributions and capture uncertainty, enhancing robustness to subtle quality differences. CoFInAl~\cite{CoFInAl} formulates AQA as a hierarchical coarse-to-fine classification problem, integrating temporal fusion and grade parsing to support fine-grained, generalizable predictions. More recently, Okamoto et al.~\cite{okamoto2024hierarchical} introduced a neuro-symbolic framework that extracts interpretable action descriptors and applies symbolic reasoning, enabling transparent and expert-aligned evaluations.

In this paper, we benchmark the main representatives of these approaches on \PS, showing that they achieve limited performance when confronted with real-world videos of procedural activities. This highlights that future research efforts should be dedicated to further developing absolute skill assessment approach in the domain of procedural video, which \PS enables.

\noindent
\textbf{Pairwise Ranking Methods}
Skill assessment has also been approached as a pairwise ranking task, where the goal is to determine which of two videos demonstrates a higher level of skill. 
The work of~\cite{doughty2018s} proposes a deep learning method to rank skill levels from videos using a novel loss function that adapts to skill differences, achieving 70–83\% accuracy across tasks like surgery and drawing, enabling automated skill assessment and video organization.
The RAAN model~\cite{doughty2019pros} assesses relative skill in long videos using a learnable temporal attention mechanisms. 
The model focuses on identifying and attending to the most skill-relevant segments to determine overall performance. RAAN is trained using video-level supervision and a rank-aware loss function.
CoRe~\cite{yu2021group} 
predicts relative quality by contrasting a target video against a reference video with comparable features. %
The Contrastive Regression framework learns quality distinctions through paired video comparisons,
while a group-aware regression tree 
is used to hierarchically regress the correct score. 
AQA-TPT~\cite{bai2022action} introduces a temporal parsing transformer that decomposes holistic video representations into temporally segmented part-level features, enabling the capture of fine-grained variations within the same action class. %
For quality scoring, the method leverages a contrastive regression approach applied to these part-based features.

We evaluate the performance of the main pairwise ranking approach on \PS, showing their limitations, when tested on real-world procedural videos and highlighting the need for future investigations in this area.

\section{The \PS Annotation Protocol}

\noindent We annotate \PS iterating through three stages, as highlighted in Figure~\ref{fig:teaser}: stage 1) video pairs are selected for labeling following a Swiss Tournament scheme; stage 2) pairwise ranking labels are crowd-sourced for the selected pairs using Amazon Mechanical Turk (AMT); stage 3) the ELO ranking system is used to convert relative pairwise rankings into absolute scores, obtaining an absolute ranking of all labeled clips.
Stages 1-3 are iterated for $R$ rounds to obtain stable absolute rankings.
\noindent \PS, as a result, provides both relative and absolute skill annotations within a unified framework, enabling flexible evaluation protocols and supporting a broader range of training objectives.
The proposed annotation approach is scalable to large datasets, robust to noisy annotations, and, requiring no expert involvement, it can be annotated through a crowd-sourced protocol, such as Amazon Mechanical Turk (AMT),
enabling efficient and reliable skill estimation across diverse procedural domains.
\noindent Below, we describe each stage of the annotation pipeline in detail.

\noindent 
\textbf{Stage 1} This stage follows a round of a Swiss Tournament, a non-elimination tournament format commonly used in popular games such as Chess,
where maintaining a large number of players throughout the tournament is desirable. In our settings video segments are the players. Video segments are paired in each round based on their current standings (e.g., absolute scores coming from a previous stage), with those having similar scores facing each other whenever possible. Unlike single-elimination tournaments, where a single loss eliminates a participant, in a Swiss Tournament every video segment competes in every round, typically lasting a predetermined number of rounds. The pairing algorithm ensures that video segment face opponents of similar skill levels as the tournament progresses, while avoiding matching two opponents multiple times in the same tournament. This format provides a fair and comprehensive assessment of video segment skill across multiple games while maximizing playing time for all participants.
We treat a pairwise comparison between two segments of the same action as a match. At the first iteration, every segment start with an equal score, while scores are gradually updated in Stage 3. The result of this stage is a set of matched video segments, for which we should obtain pairwise ranking labels. Note that, by matching segments with similar absolute scores, we can schedule for labeling only the most informative segment pairs, avoiding trivial comparisons.\\
\noindent 
\textbf{Stage 2}
we used Amazon Mechanical Turk (AMT) to label the video pairs selected in stage 1. Each annotation indicated which of the two ``opponents'' exhibited higher skill, used to refine the absolute skill score of each clip.

We designed a generalizable annotation pipeline to efficiently collect reliable pairwise skill judgments at scale. Although implemented on AMT, the methodology is platform-agnostic and can be adapted to any system supporting video playback, qualification gating, and structured task delivery. To encourage reproducibility, we will release the code and templates used in our crowdsourcing setup.

To ensure quality, workers were pre-screened based on performance history. Only contributors with a historical approval rate above 90\% and who passed a dedicated qualification test were allowed. The test consisted of gold-standard comparisons, manually curated and updated three times across stages per dataset, to evaluate the worker’s ability to reliably discern skill differences.

\begin{figure}[t]
    \centering
    \includegraphics[width=0.98\linewidth]{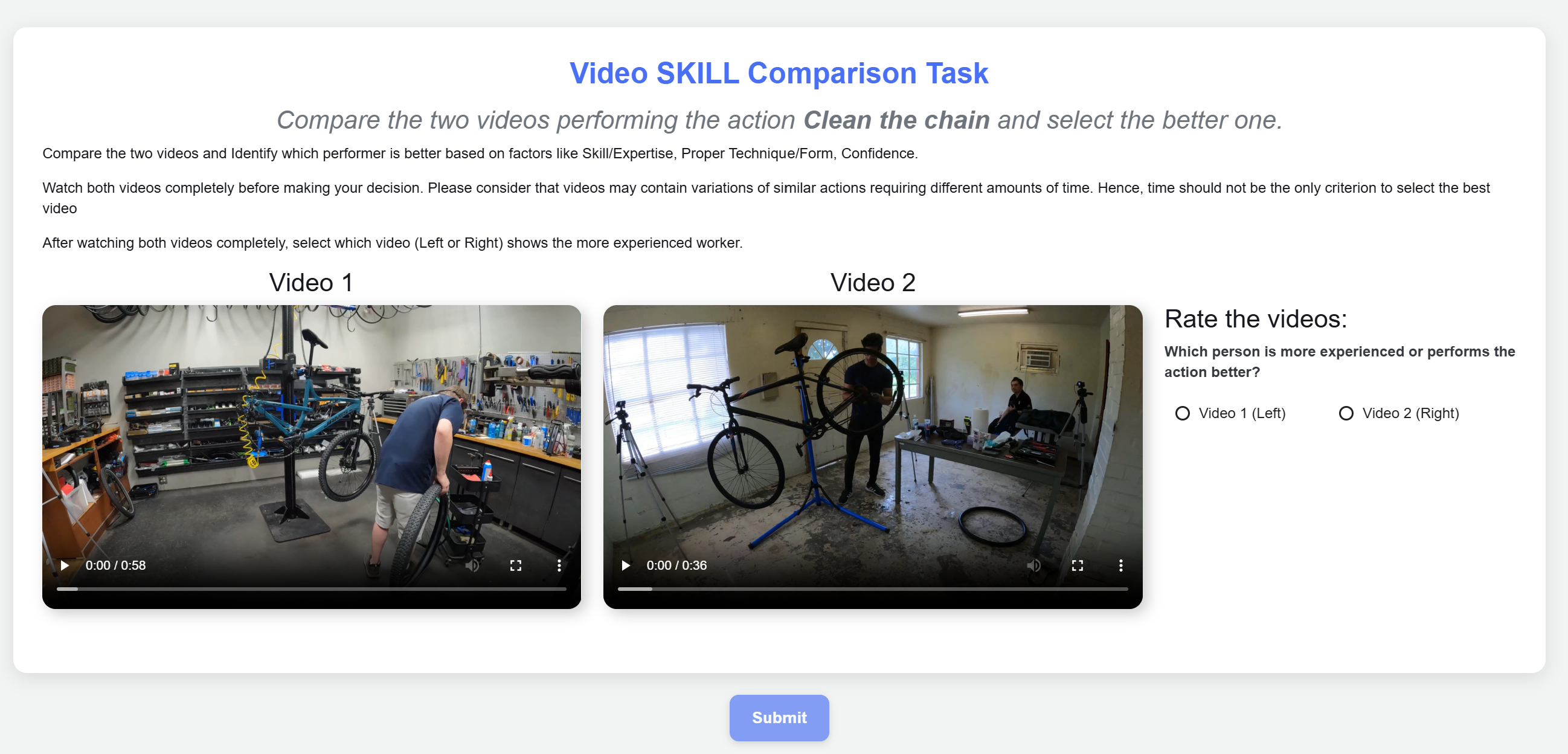}
    \caption{Screenshot of the annotation interface used in AMT. Annotators view two video segments corresponding to the same task and are asked to select which performer appears more skilled.}
    \label{fig:amt-interface-small}
\end{figure}

Each annotation task included a set of video pairs. Workers watched both videos and indicated which performer demonstrated higher skill. Instructions included positive and negative examples. The interface was designed to focus attention on the comparison itself (see Fig.~\ref{fig:amt-interface-small}). Each pair was annotated by five independent workers, with the final outcome computed by majority vote. In cases of low agreement, the pair was re-assigned to three additional workers. %
The output of this stage is a set of pairwise labels for previously matched video segments. Each label can be interpreted as the outcome of the associated match, stating which of the two segment ``wins'' the match (i.e., has the highest skill score). These labels will be used in stage 3 to obtain a consistent global ranking among segments.
Following the outlined procedure, overall, we collected 16,372 unique comparisons, annotated by 551 qualified workers, which have been used to produce the proposed \PS benchmark Dataset. See the supplementary material for further details on the AMT platform implementation and agreement rate statistics.

\noindent
\textbf{Stage 3} 
The goal of this stage is to update the absolute ranking of video segments using the the ELO rating system~\cite{elo1978rating}. 
The system was developed by physicist Arpad Elo to calculate the relative skill levels of players in zero-sum games such as Chess. In our case ELO assigns each video segment a numerical rating that increases with wins and decreases with losses. The magnitude of the rating's change is determined by the difference in ratings between opponents and the actual outcome versus the expected outcome. The expected score for a video segment is calculated using the logistic function:
\begin{equation}
    \begin{array}{r@{\hskip 1.2em}r}
    E_A = \frac{1}{1 + 10^{(R_B - R_A)/400}} &
    E_B = \frac{1}{1 + 10^{(R_A - R_B)/400}}
    \end{array}
    \label{eq:1}
\end{equation} 
where $R_A$ and $R_B$ are the current absolute ratings of video segment A and B respectively. After a match, the new rating is computed as 
\begin{equation}
    R'_A = R_A + K(S_A - E_A) 
    \label{eq:2}
\end{equation} 
where $K$ is a constant that determines the maximum possible rating change, and $S_A$ is the actual score (1 for a win, 0.5 for a draw, 0 for a loss). This self-correcting system ensures that ratings accurately reflect current playing strength.
In our setting, this method permits to assign a global expertise level to each clip, without the necessity for annotators to globally rank all the clips.
In our settings, we start from video segment pairs $(A,B)$ annotated in stage 2, where $K=1$ if $A$ is selected over $B$ and $K=0$ otherwise.
We obtain $R_A$ and $R_B$ as the absolute scores of segments $A$ and $B$ from the previous round, initialized to zero before the first round. 
Absolute scores are hence updated using \eqref{eq:1}-\eqref{eq:2}.
Once the absolute scores are updated, we can return to stage 1 to select informative segment pairs to label for pairwise ranking.

\section{The \PS Dataset}

\begin{table*}[!ht]
\centering
\setlength{\tabcolsep}{20pt}
\resizebox{0.9\textwidth}{!}{%
\begin{tabular}{lcccccc}
\toprule
& \textbf{Ikea}~\cite{ben-shabat_ikea_2023} & \textbf{Meccano}~\cite{ragusa2021meccano} & \textbf{Assembly101}~\cite{sener2022assembly101} & \textbf{Egoexo4D~\cite{grauman_ego-exo4d_2024} } & \textbf{EpicTent}~\cite{jang2019epic} & \textbf{Total} \\
\midrule
Total Clips & 160 & 80 & 560 & 191 & 144 & 1135 \\
Total Actions &10 & 5 & 35 & 12 & 9 & 71 \\
Total Time (hr) &1.28 & 1.06 & 5.49 & 4.70 & 1.59 & 14.12 \\
AVG $\pm$ STD (s) &28.88 $\pm$ 19.69 & 47.59 $\pm$ 21.45 & 35.3 $\pm$ 25.27 & 88.14 $\pm$ 90.93 & 39.71 $\pm$ 34.18 & 44.75 $\pm$ 48.46 \\
\midrule 
Train Set (\#clips) & 100 & 48 & 350 & 119 & 85 & 702 \\
Test Set (\#clips) & 40 & 19 & 140 & 48 & 36 & 283\\
Validation Set (\#clips) & 20 & 13 & 70 & 24 & 23 & 150\\
\bottomrule
\end{tabular}
}
\caption{Statistics of the \PS dataset.}
\label{tab:statistics}
\end{table*}

\noindent
\textbf{Data Selection}
\PS is based on established, public datasets of procedural videos, to which we add segment-level skill annotations following the proposed protocol. 
We chose these datasets because they provide clear examples of procedural activities where expertise is clearly assessable, and they complement each other.
Specifically, we consider the following datasets:
\begin{itemize}[noitemsep]
    \item 
\emph{EgoExo4D}~\cite{grauman_ego-exo4d_2024}: we focused on bike repair data selecting 4 scenarios for a total of 191 clips across 12 actions.
\item\emph{Meccano}~\cite{ragusa2021meccano}: we sourced segment annotations from EgoProcel~\cite{EgoProceLECCV2022} annotations - the dataset has a single scenario for a total of 80 clips across 5 actions.
\item\emph{EpicTent}~\cite{jang2019epic}: we sourced segment annotations from EgoProcel~\cite{EgoProceLECCV2022} annotations - the dataset has a single scenario for a total of 144 clips across 9 actions. 
\item\emph{IkeaASM}~\cite{ben-shabat_ikea_2023}: we selected 4 scenarios for a total of 160 clips across 10 actions.
\item\emph{Assembly101}~\cite{sener2022assembly101}: we selected 2 scenarios for a total of 560 clips across 35 actions.
\end{itemize}
\noindent
For each action, we select $16$ clips, which are labeled for pairwise and absolute skill scores using the proposed annotation protocol.
Note that, since comparing segments depicting different actions is unfeasible, we label segments from each action independently, obtaining a different absolute ranking for each action.
Absolute scores are hence converted to percentiles to be mapped to the same scale.

\begin{figure}[t]
    \centering
    \begin{subfigure}[b]{0.99\linewidth}
        \includegraphics[width=\linewidth]{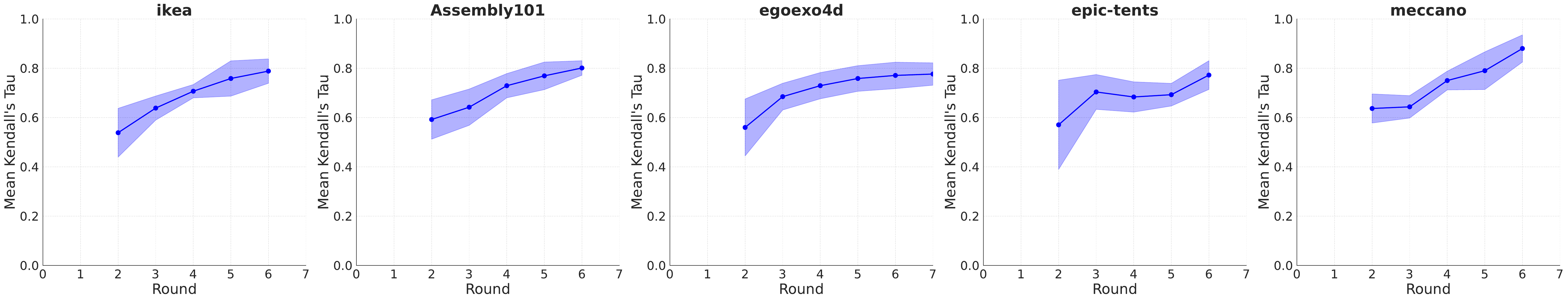}
        \caption{Kendall’s $\tau$ between consecutive rounds. A clear stabilization trend is observed for both IKEA Assembly and EgoExo4D, indicating convergence of absolute rankings over time.}
    \end{subfigure}
    \vskip\baselineskip
    \begin{subfigure}[b]{0.99\linewidth}
        \includegraphics[width=\linewidth]{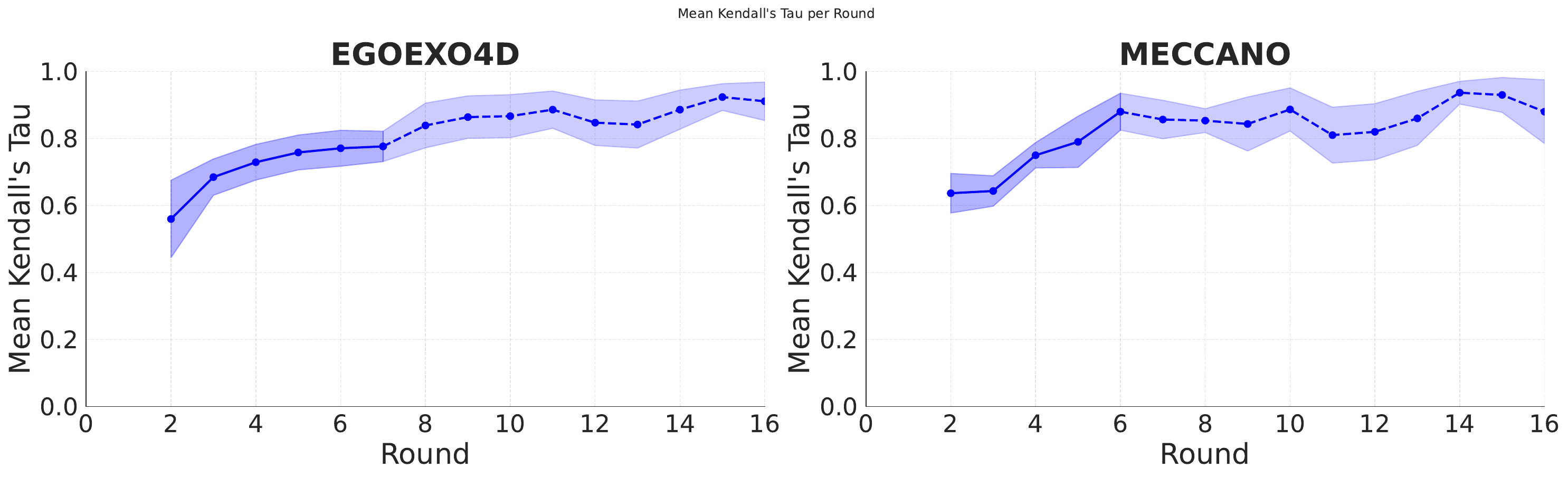}
        \caption{Comparison of ranking stability across rounds, when extending rounds beyond $6$, for Meccano and EgoExo4D. Meccano exhibits earlier convergence, while EgoExo4D stabilizes more gradually.}
    \end{subfigure}
    \vskip\baselineskip
    \caption{Ranking stability analysis across datasets. (a) Temporal evolution of Kendall’s $\tau$ across consecutive rounds. (b) Comparative round-level stability, highlighting convergence behavior. %
    }
    \label{fig:rank_stability}
\end{figure}

\noindent
\textbf{Annotation Details}
We set the number of rounds $R=6$. This number exceeds the minimum recommended number of rounds, which is usually set to the base 2 logarithm of the number of participants~\cite{fuhrlich2022improving}, which is set to $16$ segment per action in our case.
Following this scheme, we annotated approximately 40\% of all possible video segment pairings for each action. %
While more rounds and labeled pairs generally lead to finer-grained differentiation, we observe that $6$ rounds already provides quite stable results, while significantly reducing the number of comparisons.
To quantitatively assess ranking quality, we use Kendall’s tau, a rank correlation coefficient that measures the ordinal association between two variables and indicates the similarity of the orderings (rankings) of the two sets.
Figure~\ref{fig:rank_stability}a reports Kendall’s $\tau$ coefficients computed between rankings at consecutive rounds, which measures ranking consistency as annotation progresses. Datasets such as \textit{IKEA}, \textit{Assembly101} and \textit{EgoExo4D}, exhibit a clear flattening trend (with $\tau \approx 0.8$), indicating convergence toward a stable ordering. In contrast, \textit{Epic-Tents} and \textit{Meccano} display a slower upward trend, suggesting that rankings may still shift with additional annotations.

To further assess the effect of extending the annotation to more rounds, we extended beyond 6 rounds for the representative datasets, \textit{EgoExo4D} and \textit{Meccano}, to a full round-robin scheme (i.e., all possible segment pairings).
Results are shown in Figure~\ref{fig:rank_stability}b.
Coherently with our previous observation, the Kendall’s $\tau$ correlation between consecutive rounds %
stabilizes around round 6 for \textit{Meccano}. In \textit{EgoExo4D}, although rankings also appear stable around round 6, continuing the annotation process to round-robin yields a moderate further increase in Kendall’s $\tau$ (about 6\%), but also leading to a much larger number of comparisons ($\approx 60\%$ more).

\begin{figure}
    \centering
    \includegraphics[width=0.98\linewidth]{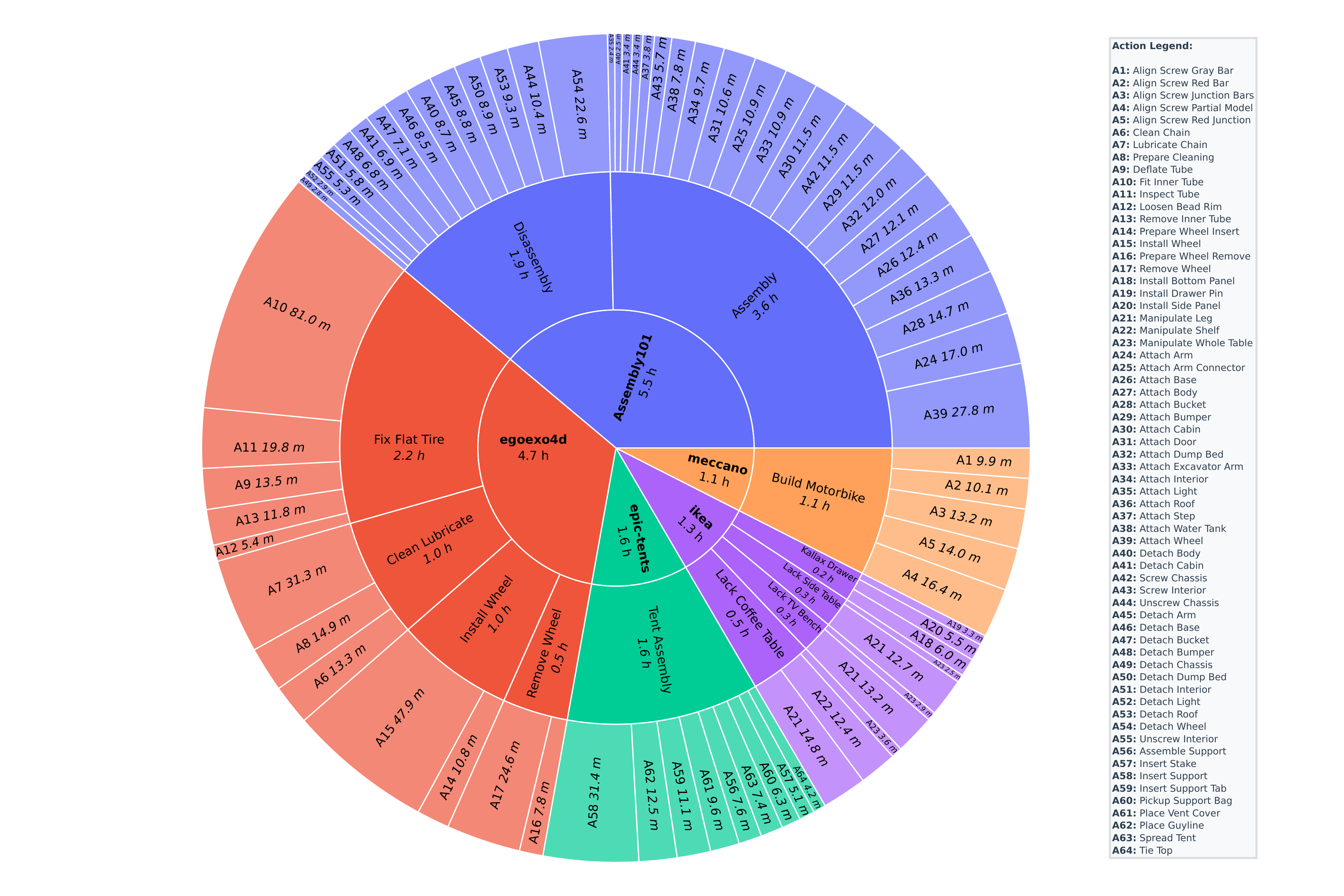}
    \caption{Illustration of the hierarchical structure of the ProSkill dataset, decomposing total annotation time by dataset, overall task type, and individual actions. Segment sizes are proportional to cumulative durations, highlighting dominant activities across different tasks. Action labels are abbreviated for readability, with a side legend offering full descriptions.}
    \label{fig:sunburst_chart}
\end{figure}

\begin{figure}
    \centering
    \includegraphics[width=0.98\linewidth]{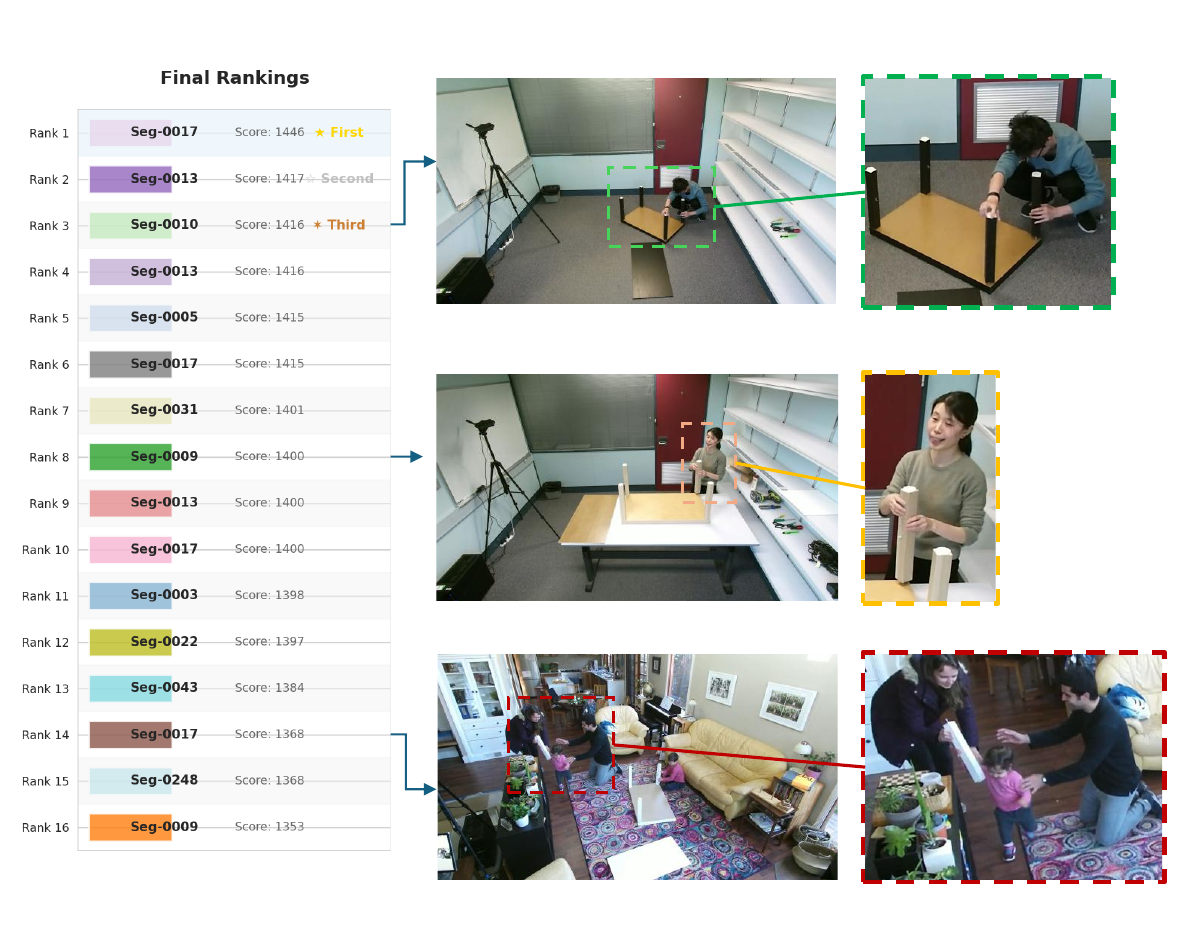}
    \caption{Example of skill ranking for the action \textit{Mount the table legs} from the IkeaASM dataset. Clips show examples of high (green), medium (yellow), and low (red) ELO-based scores, illustrating differences in efficiency, focus, and interruptions.}

    \label{fig:Qualitative}
\end{figure}

\noindent
\textbf{Data Statistics}
\PS is composed of a total of $1135$ clips and $71$ actions. This makes it a large and diverse dataset for skill assessment. Action numbers, length, dataset spits are summarized in Table~\ref{tab:statistics}. 
Figure~\ref{fig:sunburst_chart} summarizes the distribution of video length across datasets, tasks, and actions to illustrate the diversity and real-world quality of \PS.
Differently from previous datasets, we cover a wide range of activities and naturally obtain an unbalanced dataset, where the length of actions is affected by the chosen activity.
Figure~\ref{fig:Qualitative} reports ranked clips from a representative action (\textit{Mount the table legs}) in the \textit{IkeaASM} dataset. The figure showcases three clips positioned at different skill levels, top-ranked, mid-ranked, and low-ranked, according to their final ELO-based score.
The top-ranked clip features a highly committed performer who skillfully attaches two legs simultaneously, one with each hand, demonstrating both efficiency and confidence. The mid-ranked clip shows a participant performing the task slowly, occasionally pausing to speak, indicating a less focused execution. The lowest-ranked clip reveals an external disturbance, a child interfering with a table leg during the procedure, affecting the fluency of the action.
See the supplementary material for additional qualitative examples.

To ensure a robust and unbiased evaluation, the dataset was split into training, validation, and test sets at the video level. Videos were then assigned to the test, training, or validation sets such that each split contained a predefined number of clips per action, and no video appeared in more than one split. Split sizes are summarized in Table~\ref{tab:statistics}.

\section{Experiments}
\noindent
\PS enables for the first time a comprehensive benchmark of state-of-the-art approaches for video-based procedural skill assessment. 
In this section, we establish baselines to provide a clear measure of the current capabilities and limitations in the field.

\noindent
\textbf{Problem Formulation} 
We tackle skill assessment from video under two complementary setups: a \textit{regression} setting, where a model predicts a continuous score \( \hat{s}_i \) for each performance \( i \), and a \textit{pairwise ranking} setting, where the goal is to infer the relative ordering between two performances \( (i, j) \). For regression, evaluation is based on Spearman’s rank correlation coefficient \( \rho \) between predicted and ground-truth scores, assessing the preservation of relative skill order. In the pairwise setting, we compute the winner prediction accuracy, i.e., the proportion of correctly predicted preferences.

\noindent
\textbf{Baselines}
\label{sec:benchmarks}
\noindent
For Global Ranking, we compare three representative approaches: USDL~\cite{usdl}, DAE-AQA~\cite{10.1007/dae-aqa}, and CoFInAl~\cite{CoFInAl}. For USDL~\cite{usdl}, we adopt the single-judge variant, as our scores, although derived from multiple annotators, are aggregated using the ELO algorithm and treated as a single ground-truth value. Both USDL~\cite{usdl} and DAE-AQA~\cite{10.1007/dae-aqa} operate on a single [1, 1024] feature vector per video, obtained by aggregating features extracted with a temporal sliding window. During training, we sample one of the $T$ available representations for each clip. In contrast, CoFInAl~\cite{CoFInAl} is designed to work with multiple temporal representations and processes four feature vectors per clip. 
For Pairwise Ranking we evaluate the following three representative approaches: RAAN~\cite{doughty2019pros}, AQA-TPT~\cite{bai2022action}, and CoRe~\cite{yu2021group}.
\noindent The first baseline, RAAN~\cite{doughty2019pros}, operates by comparing two input videos and predicting which one demonstrates higher skill. Importantly, both videos are sampled from the test set and were never seen during training, making this a true generalization test. RAAN~\cite{doughty2019pros} outputs a binary decision indicating the ``winner" between the two clips, based solely on relative performance.
In contrast, the other two baselines, AQA-TPT~\cite{bai2022action} and CoRe~\cite{yu2021group}, perform evaluation using a reference-based approach. Each takes as input a test video and compares it against a training video depicting the same action. They estimate a score delta, reflecting how the performance in the test video deviates from that in the training example. While this method provides fine-grained feedback, it relies on access to training examples during inference, limiting its generalization compared to RAAN~\cite{doughty2019pros}.

\noindent
\textbf{Implementation Details}
The considered baselines involve several hyperparameters that strongly influence performance.
For fair comparisons, we adopt a unified feature extraction pipeline and tune hyperparameters as detailed below.
Each video is split into overlapping 16-frame clips, resized to 256 px, center-cropped to 224×224, and encoded with either I3D~\cite{carreira2017quo} or VideoMAE~\cite{tong2022videomae}, yielding 1024-D features.
Unlike prior datasets, \PS covers a wide range of actions, making per-action models impractical.
We therefore train a single model across all actions for each subset (Meccano, Egoexo4D, Ikea, Assembly101, EpicTent).
During training, to maximize validation performance, we optimize three hyperparameters per model: score normalization (except for RAAN), noise augmentation, and learning rate.
Full grid-search details are reported in the supplementary material.

\noindent
\textbf{Global Ranking Results}
Table~\ref{tab:global_ranking_rho} compares Spearman’s $\rho$ across five datasets using global ranking methods (USDL, DAE-AQA, CoFInAl) and pairwise regression models (AQA-TPT, CoRe).
Global ranking models generally outperform pairwise ones, suggesting that leveraging global context improves skill assessment. \textsc{CoFInAl} achieves the highest correlations overall, notably on \textit{Meccano} ($\rho=0.59$), \textit{Ikea}, and \textit{EpicTent}, demonstrating strong temporal modeling. \textsc{USDL} shows consistent results, especially on \textit{Egoexo4D}, while \textsc{DAE-AQA} is less stable. Among pairwise models, \textsc{AQA-TPT} performs moderately on \textit{Egoexo4D} but poorly elsewhere; \textsc{CoRe} has mixed results with some strengths on \textit{Assembly101} and \textit{Egoexo4D}.
VideoMAE features generally lead to better correlations than I3D, except on \textit{Assembly101} where I3D is slightly better, indicating domain-specific differences.
Dataset-wise, \textit{Meccano} is the easiest to predict, likely due to its controlled setting, while \textit{Assembly101} is the most challenging due to high variability and noisy annotations. Overall, results confirm that skill assessment in complex, multi-action tasks remains a challenging problem requiring improved models and features.

\begin{table}[t]
\centering
\resizebox{\columnwidth}{!}{%
\begin{tabular}{llccccc}
\toprule
\textbf{Method} & \textbf{Features} &  \textbf{Ikea}~\cite{ben-shabat_ikea_2023} & \textbf{Meccano}~\cite{ragusa2021meccano} & \textbf{Assembly101}~\cite{sener2022assembly101} & \textbf{Egoexo4D~\cite{grauman_ego-exo4d_2024} } & \textbf{EpicTent}~\cite{jang2019epic} \\
\midrule

\multirow{2}{*}{\textsc{USDL~\cite{usdl}}} 
  & \textcolor{gray}{I3D}        &0.12  &0.38  &    0.12   & 0.33   &   0.17    \\
  & \textcolor{gray}{VideoMAE}   & 0.19 & \textbf{0.43} &    0.13   & \underline{0.39}    &   0.23    \\
    \arrayrulecolor{gray!50}\midrule
\multirow{2}{*}{\textsc{DAE-AQA~\cite{10.1007/dae-aqa}}} 
  & \textcolor{gray}{I3D}         & 0.20 & 0.24 &  0.20     & 0.16    &  0.23     \\
  & \textcolor{gray}{VideoMAE}   & 0.10 & \textbf{0.42} &   0.03    &    0.33 &    \underline{0.26}   \\

\arrayrulecolor{gray!50}\midrule
\multirow{2}{*}{\textsc{CoFInAl~\cite{CoFInAl}}} 
  & \textcolor{gray}{I3D}         &\underline{0.26}  & \textbf{\underline{0.59}} &    0.14   &  0.20   &   0.21    \\
  & \textcolor{gray}{VideoMAE}    & \underline{0.26} & 0.31 &    0.11   &  0.28   &   0.23    \\

\arrayrulecolor{black} %
\midrule
\rowcolor{gray!10}
& \textcolor{gray}{I3D} & 0.14 & 0.12 & -0.02 & 0.17 & -0.01\\
\rowcolor{gray!10}
\multirow{-2}{*}{\textsc{AQA-TPT~\cite{bai2022action}}}  & \textcolor{gray}{VideoMAE} & 0.21 & 0.35 & 0.15 & \textbf{0.36} & -0.01 \\
\arrayrulecolor{gray!50}\midrule
\rowcolor{gray!10}
 & \textcolor{gray}{I3D} & 0.22 & -0.12 & \underline{0.22} & 0.33 & 0.04\\
\rowcolor{gray!10}
\multirow{-2}{*}{\textsc{CoRe~\cite{yu2021group}}} & \textcolor{gray}{VideoMAE} & 0.19 & 0.24 & 0.06 & \textbf{0.35} & 0.12   \\

\arrayrulecolor{black}\midrule
\rowcolor{blue!10}
  & \textcolor{gray}{I3D}         &0.20 &   0.24  &      0.13    & 0.24   &  0.13   \\
 \rowcolor{blue!10} 
\multirow{-2}{*}{\textsc{Average}}   & \textcolor{gray}{VideoMAE}    & 0.20  &  0.35    &    0.10  &   0.34  &   0.17   \\
\arrayrulecolor{black} \bottomrule
\end{tabular}
}
\caption{Spearman’s $\rho$ for global ranking. A single model is trained for all the actions. Evaluated with I3D and VideoMAE features. White backgroud rows use absolute ranking, gray background uses pairwise ranking. Bold best per method, Underline best per Dataset.}
\label{tab:global_ranking_rho}
\end{table}

\noindent
\textbf{Single-Action Model vs Multi-Action Model}
Table~\ref{tab:global_ranking_rho_sub} compares USDL when training a unified model for all actions in a dataset or a single model per action, a setup commonly used in AQA approaches.
Results show that unified models consistently outperform per-action models, which often suffer from poor or even negative correlations due to limited supervision at the action level. 
This highlights how in a realistic scenario, as the one offered by \PS, single-action models obtain limited performance.

\begin{table}[t]
\centering
\resizebox{\columnwidth}{!}{%
\begin{tabular}{llccccc}
\toprule
\textbf{Method} & \textbf{Features} &  \textbf{Ikea}~\cite{ben-shabat_ikea_2023} & \textbf{Meccano}~\cite{ragusa2021meccano} & \textbf{Assembly101}~\cite{sener2022assembly101} & \textbf{Egoexo4D~\cite{grauman_ego-exo4d_2024} } & \textbf{EpicTent}~\cite{jang2019epic} \\
\midrule
\multirow{2}{*}{\textsc{USDL~\cite{usdl}}} 
  & \textcolor{gray}{I3D}        &0.12  &0.38  &    0.12   & 0.33   &   0.17    \\
  & \textcolor{gray}{VideoMAE}   & 0.19 & 0.43 &    \underline{0.13}   & 0.39    &   \underline{0.23}    \\
\midrule
\multirow{2}{*}{\textsc{USDL - Single~\cite{usdl}}} 
  & \textcolor{gray}{I3D}        & -0.18 & 0.09 & -0.09 & 0.19  & 0.17 \\
  & \textcolor{gray}{VideoMAE}   & 0.08 & 0.01 & -0.31 &  0.04 & 0.06   \\

\arrayrulecolor{black} %
\bottomrule
\end{tabular}
}
\caption{Spearman’s $\rho$ for global ranking training a model for each sub-actions.}
\label{tab:global_ranking_rho_sub}
\end{table}

\noindent
\textbf{Textual Grounding}
While training action-specific models is unfeasible in real settings, the knowledge of the action to be assessed may still be available at test time, hence constituting a useful prior for skill prediction.
To assess this hypothesis, we propose to condition the unified model on a textual representation of the action (e.g., the action name) to enable it to adapt its scoring to the specific semantic context. 
Specifically, we conduct experiments with USDL by combining video features with textual embeddings extracted using a MiniLM pre-trained model~\cite{wang2020minilmdeepselfattentiondistillation}.
As shown in Table~\ref{tab:global_ranking_rho_sub_2}, conditioning the unified model on action descriptions provides modest but consistent improvements in several datasets, suggesting that textual grounding can offer helpful contextual information.

\begin{table}[t]
\centering
\resizebox{\columnwidth}{!}{%
\begin{tabular}{llccccc}
\toprule
\textbf{Method} & \textbf{Features} &  \textbf{Ikea}~\cite{ben-shabat_ikea_2023} & \textbf{Meccano}~\cite{ragusa2021meccano} & \textbf{Assembly101}~\cite{sener2022assembly101} & \textbf{Egoexo4D~\cite{grauman_ego-exo4d_2024} } & \textbf{EpicTent}~\cite{jang2019epic} \\

\midrule
\multirow{2}{*}{\textsc{USDL~\cite{usdl}}} 
  & \textcolor{gray}{I3D}        &0.19  &0.38  &    0.12   & 0.33   &   0.17    \\
  & \textcolor{gray}{VideoMAE}   & 0.22 & 0.43 &    \underline{0.13}   & 0.39    &   \underline{0.23}    \\
\midrule
\multirow{2}{*}{\textsc{USDL~\cite{usdl} + Grounding}} 
  & \textcolor{gray}{I3D+MiniLM}        & 0.24 &0.36  & 0.12    & 0.33   & 0.20    \\
  & \textcolor{gray}{VideoMAE+MiniLM}   & \underline{0.27} &\underline{0.50}     & \underline{0.13}  & \underline{0.41}  &  0.18   \\

\arrayrulecolor{black} %
\bottomrule
\end{tabular}
}
\caption{Spearman’s $\rho$ for global ranking. A single model is trained for all the actions considering also action text embeddings. Evaluated with I3D and VideoMAE features. Underline best per Dataset.}
\label{tab:global_ranking_rho_sub_2}
\end{table}

\begin{table}[t]
\centering
\resizebox{\columnwidth}{!}{%
\begin{tabular}{llccccc}
\toprule
\textbf{Method} & \textbf{Features} &  \textbf{Ikea}~\cite{ben-shabat_ikea_2023} & \textbf{Meccano}~\cite{ragusa2021meccano} & \textbf{Assembly101}~\cite{sener2022assembly101} & \textbf{Egoexo4D~\cite{grauman_ego-exo4d_2024} } & \textbf{EpicTent}~\cite{jang2019epic} \\
\midrule
\multirow{2}{*}{\textsc{AQA-TPT~\cite{bai2022action}}} & \textcolor{gray}{I3D} & 0.71 & 0.66 & 0.53 & 0.72 & 0.70 \\
 & \textcolor{gray}{VideoMAE} & \underline{0.73} & \underline{0.74} & \underline{0.70} & \textbf{\underline{0.79}} & \underline{0.76} \\
\arrayrulecolor{gray!50}\midrule
\multirow{2}{*}{\textsc{CoRe~\cite{yu2021group}}} & \textcolor{gray}{I3D} & \underline{0.73} & 0.62 & 0.69 & \textbf{0.76} & 0.67 \\
 & \textcolor{gray}{VideoMAE} & 0.70 & 0.73 & 0.69 & 0.75 & 0.75\\
\arrayrulecolor{gray!50}\midrule
\multirow{2}{*}{\textsc{RAAN~\cite{doughty2019pros}}} & \textcolor{gray}{I3D} & 0.45 & 0.52 & 0.51 & 0.58 & 0.52 \\
 & \textcolor{gray}{VideoMAE} & \textbf{0.68} & 0.52 & 0.51 & 0.46 & 0.57 \\
\arrayrulecolor{black}\midrule
\rowcolor{blue!10}
 & \textcolor{gray}{I3D} & 0.63 & 0.60 & 0.57  & 0.68 & 0.63 \\
\rowcolor{blue!10}
\multirow{-2}{*}{\textsc{Average}} & \textcolor{gray}{VideoMAE} & 0.70 & 0.66 & 0.63  & 0.66 &  0.69\\  
\bottomrule
\end{tabular}
}
\caption{Pairwise ranking accuracy. In AQA-TPT~\cite{bai2022action} and CoRe~\cite{yu2021group} one video is from test the other is an example from train, in RAAN~\cite{doughty2019pros} both are from test. Bold best per method, Underline best per Dataset.}
\label{tab:pairwise_ranking_acc}
\end{table}

\noindent
\textbf{Pairwise Ranking Results}
\label{sec:pairwise_ranking}Table~\ref{tab:pairwise_ranking_acc} shows pairwise ranking accuracy results across datasets and features. On average VideoMAE has better results over I3D across all subsets, except \textit{Ego-Exo4D}, according to this evaluation measure. The best result of all experiments is obtained using AQA-TPT with VideoMAE on \textit{Ego-Exo4D} $0.79$, whereas the worst result is obtained by RAAN using I3D on \textit{Ikea} subset $0.45$. \textit{Assembly101} confirms to be the most challenging subset also in the pairwise ranking scenario: averaging across methods and features it reach an accuracy of $0.60$, just above chance level. \textit{Meccano} is challenging as well, reaching low average scores of $0.63$ and $0.6$ depending on the choice of features.
Overall, these results confirm that performing skill assessment in procedural video is challenging, suggesting that \PS is a useful resource to advance the field.

\section{Conclusions}
\noindent We introduced  \PS, the first dataset specifically designed for action-level skill assessment in procedural tasks. Differently from previous datasets, \PS offers both pairwise and absolute ranking annotations, collected through a scalable ELO-based annotation protocol that combines the efficiency of crowdsourcing with the reliability of tournament-style comparisons.
Our experiments show that existing state-of-the-art methods, originally developed for more constrained or domain-specific settings, tend to underperform on \PS, particularly in global ranking tasks. Pairwise methods generally yield better results, but the overall performance indicates that skill assessment in procedural videos remains a challenging and largely unsolved problem. \PS serves as a valuable benchmark for the evaluation of models under realistic conditions, and for investigation of the numerous open questions in human performance analysis. We will make the dataset and annotation protocol publicly available for academic research. %

\section*{Acknowledgements} 
This research is supported by Toyota Motor Europe, Next Vision s.r.l. and by the project Future Artificial Intelligence Research (FAIR) – PNRR MUR Cod. PE0000013 - CUP: E63C22001940006.

{\small
\bibliographystyle{ieee_fullname}
\bibliography{egbib}
}

\begin{appendices}

\section{Amazon Mechanical Turk}
\noindent This section will detail the design and implementation of our annotation tasks on Amazon Mechanical Turk (AMT). We describe the user interface shown to workers, the iterative round-based annotation workflow, quality control mechanisms including qualification and inter-HIT tests, and the technical tools used to automate task management and data processing.

\subsection{User Interface}
\begin{figure*}[t]
    \centering
    \includegraphics[width=1\linewidth]{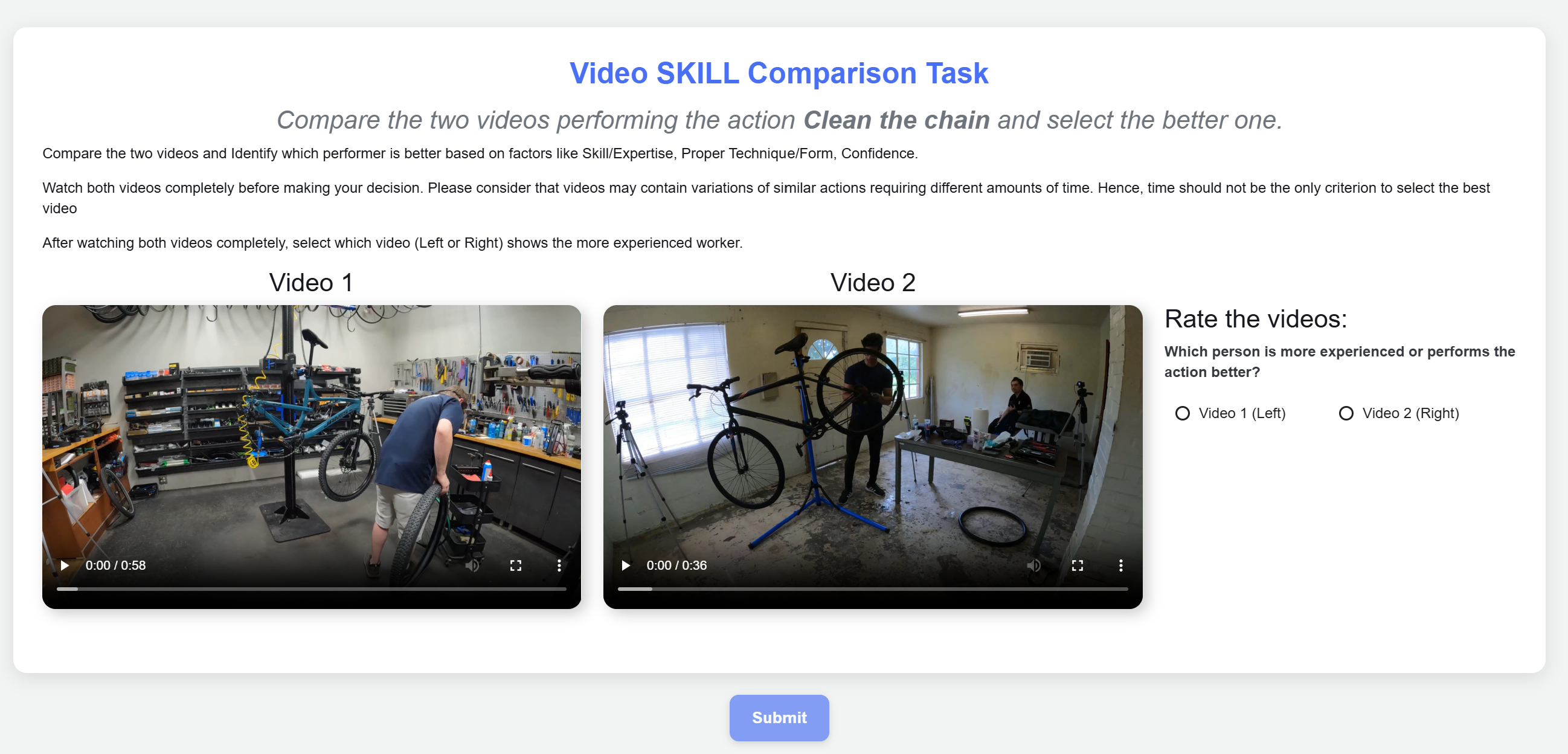}
    \caption{Screenshot of the annotation interface used in AMT. Annotators view two video segments corresponding to the same task and are asked to select which performer appears more skilled.}
    \label{fig:amt-interface}
\end{figure*}

\noindent Each HIT on AMT displays a pair of videos side by side, both showing the same sub-action (e.g., \textit{Clean the chain}). Annotators are instructed to compare the two performances and select the individual who demonstrates superior skill. See Figure~\ref{fig:amt-interface} for an example of the annotation interface.

\textbf{Prompt shown to workers:}
\begin{quote}
\emph{Compare the two videos performing the action “Clean the chain” and select the better one. Identify which performer is better based on factors like Skill/Expertise, Proper Technique/Form, Confidence.}
\end{quote}

\textbf{Instructions summary:}
\begin{itemize}
\item Workers must \textbf{watch both videos completely}.
\item They are instructed not to base their judgment solely on execution speed.
\item Emphasis is placed on evaluating \emph{technique}, \emph{confidence}, and \emph{correctness}.
\item After watching, workers select one of:
\begin{itemize}
    \item Video 1 (Left)
    \item Video 2 (Right)
\end{itemize}
\end{itemize}

\begin{figure*}[t]
    \centering
     \includegraphics[width=\linewidth]{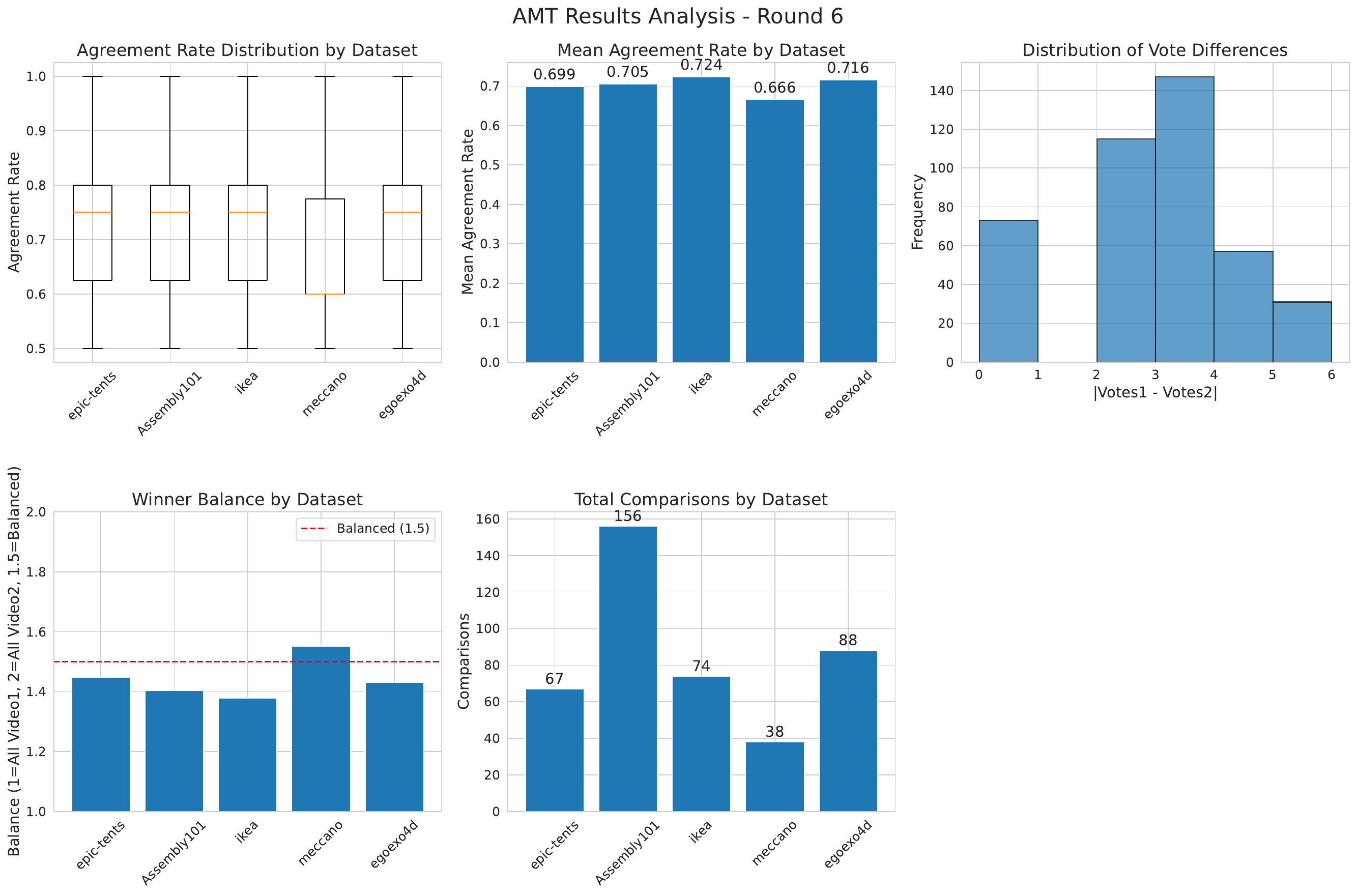}
    \caption{Analysis of AMT annotations across datasets. From left to right and top to bottom: (1) Agreement Rate Distribution, (2) Mean Agreement Rate, (3) Distribution of Vote Differences, (4) Winner Balance, (5) Total Comparisons.}
    \label{fig:amt-analysis}
\end{figure*}

\subsection{Task Design and Implementation}

\noindent The annotation workflow was organized into iterative rounds. At the end of each round, we extracted pairs of videos based on the updated global ranking to generate the next set of HITs. Each pair consisted of two video URLs hosted on our own servers, which were embedded within the AMT interface to provide seamless playback and consistent presentation.

\noindent Workers were asked to compare each pair and select which video demonstrated better skill, or indicate if there was no noticeable difference. After completing the annotations for a round, the global ranking was updated according to the collected preferences, and new pairs were generated for the subsequent round accordingly.

\noindent The user interface was developed using HTML, CSS, and JavaScript embedded within AMT’s custom templates. To ensure data quality, a qualification test using gold-standard pairs with known correct answers was required before workers could participate. Additionally, inter-HIT quality controls were implemented by randomly inserting gold-standard pairs within HITs during the annotation process, allowing real-time monitoring of worker reliability and suspension of underperforming annotators.

\noindent Task management, including HIT creation, publication, and results retrieval, was automated through Python scripts using the AWS SDK for Python (\texttt{boto3}). These scripts enabled efficient batch processing and streamlined data handling.

\noindent Preprocessing scripts prepared video clips for annotation, while aggregation pipelines combined pairwise preferences into robust global rankings. All source code, including frontend templates and backend scripts, was maintained under version control in a dedicated Git repository to ensure reproducibility and ease of maintenance.

\subsection{Annotation Statistics}

\noindent We conducted a detailed analysis of the collected annotations across five datasets and present the results below.

\paragraph{Agreement Rate Distribution by Dataset}
The top-left boxplot in Fig.~\ref{fig:amt-analysis} illustrates the distribution of agreement rates among annotators for each dataset. Across the five datasets, agreement rates generally cluster around 0.7, with some variability. IKEA and EgoExo4D tend to have slightly higher median agreement, reflecting more consistent consensus among workers. Epic-Tents and Meccano show wider variability, indicating some tasks were more ambiguous or difficult to judge consistently. This plot helps visualize how reliably annotators agree on skill assessments depending on the dataset context.

\paragraph{Mean Agreement Rate by Dataset}
The top-center bar chart in Fig.~\ref{fig:amt-analysis} shows the average agreement rate for each dataset, summarizing annotator consensus in a single metric. IKEA scores the highest mean agreement (0.724), followed closely by EgoExo4D (0.716) and Assembly101 (0.705). Meccano and Epic-Tents have slightly lower means around 0.666 and 0.699 respectively. These differences may relate to task complexity, video quality, or inherent ambiguity in skill evaluation across different procedural tasks.

\paragraph{Distribution of Vote Differences}
The top-right histogram in Fig.~\ref{fig:amt-analysis} represents how many comparisons exhibit various levels of vote difference — the absolute difference in votes between the two videos compared. Most comparisons have vote differences centered around 2 or 3, indicating moderate consensus among annotators. Few comparisons have very low (0 or 1) or very high (5+) vote differences, highlighting that while some pairs are clear winners, others are more contested or balanced in skill demonstration.

\paragraph{Winner Balance by Dataset}
The bottom-left winner balance plot in Fig.~\ref{fig:amt-analysis} visualizes how wins are distributed between the two videos compared within each dataset. Instead of the previous ratio, the balance is now expressed on a scale from 1 to 2, where a value of 1 means all votes favored the first video, 2 means all votes favored the second video, and 1.5 represents a perfectly balanced split. Datasets like Epic-Tents, Meccano, and EgoExo show values close to 1.5, indicating relatively even competition between videos. Conversely, IKEA and Assembly101 display values skewed closer to 1, suggesting a tendency for the first video to be favored more often, which may reflect stronger or more consistent skill differences in these datasets.

\paragraph{Total Comparisons by Dataset}
The bottom-center bar plot in Fig.~\ref{fig:amt-analysis} reports the total number of pairwise comparisons collected for each dataset in round 6. Assembly101 contributes the majority with 156 comparisons, reflecting its larger dataset or annotation emphasis. IKEA and EgoExo4D follow with 74 and 88 comparisons respectively, while Epic-Tents and Meccano have fewer at 67 and 38. This distribution indicates the relative annotation effort and data availability across procedural video sources. Overall, we collected 16,372 unique comparisons, annotated by 551 qualified workers.

\subsection{Labeling Time Analysis}
\label{sec:labeling-time}

\noindent In this section, we analyze the time required by annotators to complete each HIT. Understanding annotation duration provides insight into the complexity and cognitive load of the skill assessment task. To gain insights into the annotation effort required per action, we analyzed the average time spent by workers to complete each HIT, normalized by the average video clip length of the corresponding dataset. This provides a scale-invariant measure of annotation time, accounting for differences in video duration across datasets. We further rescaled these values between 0 and 1 for comparability.

\begin{figure*}[t]
    \centering
    \includegraphics[width=1\linewidth]{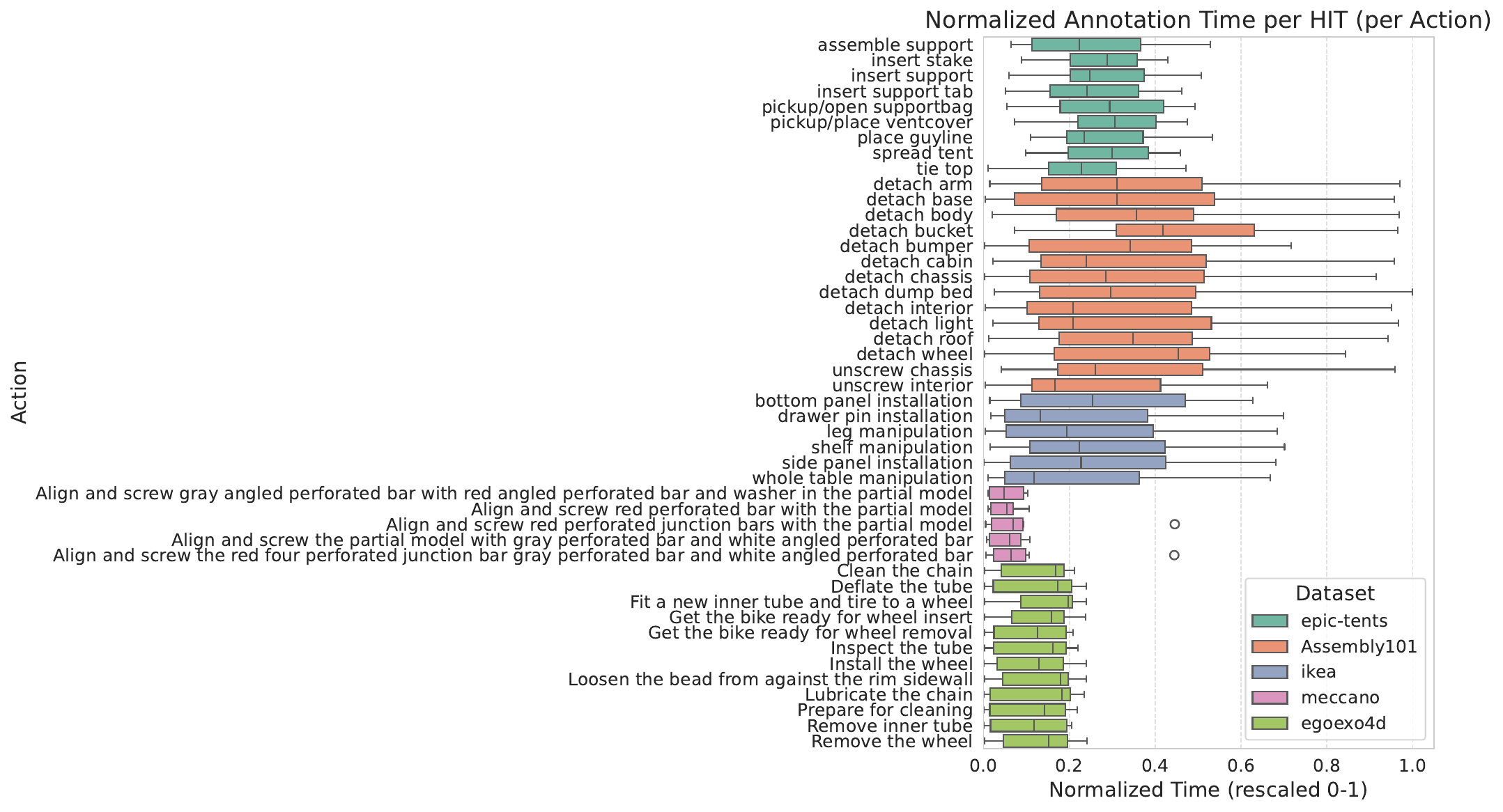}
    \caption{Distribution of normalized annotation time per HIT (rescaled to [0, 1]) across actions and datasets.}
    \label{fig:amt-time-boxplot}
\end{figure*}

\noindent As shown in Figure~\ref{fig:amt-time-boxplot}, the normalized annotation time varies considerably depending on the dataset and the complexity of the action. We report below the mean and standard deviation of normalized time for each action. In general, actions from the \textit{Egoexo4D} dataset were annotated more quickly (e.g., \textit{Remove inner tube}: $0.108 \pm 0.083$), likely due to higher visual clarity or simpler temporal dynamics. Conversely, actions in \textit{Assembly101} show higher values, such as \textit{detach bucket} ($0.452 \pm 0.244$), reflecting their multistep nature and complex object manipulation.

\noindent Annotations for \textit{Meccano} are particularly efficient, with most actions taking less than 0.13 normalized time units, suggesting clear, concise clips and high inter-rater consistency. For \textit{EPIC-Tents}, most actions range between $0.24$ and $0.30$, indicating moderate complexity and uniform clip duration. \textit{IKEA} annotations fall in a similar range but with slightly higher variance, particularly for actions involving furniture manipulation such as \textit{bottom panel installation} and \textit{shelf manipulation}.

\noindent These findings confirm that task complexity, action type, and visual clarity play a key role in determining annotation time. Importantly, the normalized time metric can be used to estimate annotation cost and design future annotation tasks with balanced load per HIT.

\noindent To further interpret these findings, we administered a short feedback questionnaire at the end of the pilot. Two main observations emerged:
First, the notions of skill, technique, and confidence were perceived as highly correlated and often difficult to distinguish. Second, annotators reported that they rarely selected a single factor, instead opting for multiple checkboxes to reflect an intertwined rationale.

\noindent Based on these insights, we chose to omit the justification checkboxes in the final annotation protocol used in \textsc{ProSkill}. While well-intentioned, these labels introduced ambiguity and cognitive overhead, and their high co-occurrence suggested they did not yield clearly disentangled supervisory signals.

\section{Swiss Tournament Round Generation}
\label{sec:swiss-generation}

\noindent To efficiently collect pairwise comparisons without exhaustively evaluating all video pairs, we adopt a Swiss-style tournament protocol that operates in rounds. After each round $r$, a global ranking is updated and used to generate the set of comparisons for the next round $r{+}1$. 

\noindent Let $\mathcal{S} = \{s_1, s_2, \dots, s_n\}$ be the set of video segments to be ranked within a sub-task. At round $r$, we maintain:
\begin{itemize}
    \item A set of match outcomes $\mathcal{M}^{(r)} = \{(s_i, s_j, y_{ij})\}$, where $y_{ij} = 1$ if $s_i$ is preferred over $s_j$, $0$ otherwise, in draw, i.e. no agreements on AMT labeling, 0.5.
    \item A ranking function $\pi^{(r)}: \mathcal{S} \rightarrow \mathbb{R}$ that assigns a score to each segment based on outcomes so far.
\end{itemize}

\noindent We estimate $\pi^{(r)}$ using a combination of ELO rating and a Swiss tournament point system. The ELO score for segment $s_i$ after round $r$ is recursively updated using:
\[
\text{Elo}_i^{(r+1)} = \text{Elo}_i^{(r)} + K \cdot \left( y_{ij} - p_{ij} \right),
\]
where 
\[
p_{ij} = \frac{1}{1 + 10^{(\text{Elo}_j^{(r)} - \text{Elo}_i^{(r)}) / 400}}
\]
is the expected probability of $s_i$ winning against $s_j$, and $K$ is a constant determining the learning rate.

Each new round selects a set of candidate pairs $\mathcal{C}^{(r+1)} \subset \mathcal{S} \times \mathcal{S}$ by:
\begin{align}
\mathcal{C}^{(r+1)} = \big\{ (s_i, s_j) \;\big|\; 
& |\pi^{(r)}(s_i) - \pi^{(r)}(s_j)| \text{ is minimal,} \notag \\
& (s_i, s_j) \notin \mathcal{M}^{(\leq r)} \big\}
\end{align}
i.e., we pair segments with similar scores that have not been compared yet.

\section{Hyperparameter optimization}
 
Table~\ref{tab:usdl_gridsearch_styled} summarizes the impact of three key hyperparameters, learning rate (\(\theta\)), augmentation noise standard deviation (\(\sigma\)), and score normalization factor (\(\tau\)),on the USDL baseline’s performance across multiple datasets and feature types (I3D and VideoMAE). Each row aggregates results by varying a single hyperparameter while averaging over the others, reporting both mean and best Spearman’s correlation coefficients on validation and test splits. The results reveal several insights: First, the optimal hyperparameters differ across datasets, reflecting their distinct data distributions and complexity. For example, Meccano and IKEA tend to favor moderate learning rates around 0.004–0.01 and normalization factors near 20–25, whereas datasets like EpicTents show greater sensitivity to augmentation noise, which can boost generalization on test data. Second, VideoMAE features consistently yield higher Spearman correlations than I3D, especially on more challenging datasets like EgoExo4D and EpicTents, suggesting stronger representation power for skill-related information. Third, augmentation noise and normalization play a crucial role in improving robustness and ranking consistency, sometimes more so than fine-tuning the learning rate.

\begin{table*}[ht]
\centering
\small
\setlength{\tabcolsep}{10pt}
\renewcommand{\arraystretch}{1.15}
\captionsetup{skip=5pt}

\rowcolors{1}{gray!3}{white}

\begin{tabular}{
  >{\columncolor{lightgray}\raggedright\arraybackslash}p{2.7cm}
  >{\raggedright\arraybackslash}l
  >{\raggedright\arraybackslash}l 
  >{\centering\arraybackslash}c
  >{\centering\arraybackslash}c
  >{\centering\arraybackslash}c
  >{\centering\arraybackslash}c
  >{\centering\arraybackslash}c
  >{\centering\arraybackslash}c
  >{\centering\arraybackslash}c
}
\toprule
\textbf{Dataset} & \textbf{Feature} & \textbf{Aggregated by} & \(\theta\) & \(\sigma\) & \(\tau\) & \(\rho_{val}^{mean}\) & \(\rho_{val}^{best}\) & \(\rho_{test}^{mean}\) & \(\rho_{test}^{best}\) \\
\midrule

Assembly101 & I3D       & lr        & 0.003 & 0.047 & 20.5 & {\textbf{0.39}} & 0.48 & 0.14 & 0.20 \\
Assembly101 & I3D       & noise     & 0.004 & 0.000 & 20.5 & 0.35 & 0.47 & 0.12 & 0.24 \\
Assembly101 & I3D       & norm      & 0.005 & 0.045 & 25.0 & 0.34 & {\textbf{0.51}} & 0.11 & {0.26} \\
Assembly101 & VideoMAE  & lr        & 0.010 & 0.047 & 20.5 & 0.21 & 0.39 & {\textbf{0.15}} & \textbf{0.27} \\
Assembly101 & VideoMAE  & noise     & 0.004 & 0.100 & 20.5 & 0.22 & 0.37 & 0.13 & 0.22 \\
Assembly101 & VideoMAE  & norm      & 0.007 & 0.044 & 16.0 & 0.18 & 0.38 & 0.12 & 0.24 \\

\midrule

Meccano & I3D       & lr        & 0.010 & 0.056 & 23.5 & 0.57 & 0.77 & \textbf{0.35} & 0.64 \\
Meccano & I3D       & noise     & 0.004 & 0.004 & 20.9 & 0.60 & 0.77 & \textbf{0.35} & 0.64 \\
Meccano & I3D       & norm      & 0.004 & 0.047 & 25.0 & {0.60} & {0.80} & 0.32 & 0.64 \\
Meccano & VideoMAE  & lr        & 0.001 & 0.047 & 20.5 & \textbf{0.61} & \textbf{0.84} & 0.30 & 0.56 \\
Meccano & VideoMAE  & noise     & 0.004 & 0.005 & 20.5 & 0.48 & 0.83 & 0.28 & 0.62 \\
Meccano & VideoMAE  & norm      & 0.004 & 0.047 & 25.0 & 0.49 &\textbf{0.84} & 0.24 & {\textbf{0.65}} \\

\midrule

EpicTents & I3D       & lr        & 0.001 & 0.035 & 19.4 & 0.05 & 0.45 & 0.13 & 0.21 \\
EpicTents & I3D       & noise     & 0.004 & 0.005 & 20.5 & 0.19 & 0.58 & 0.10 & 0.22 \\
EpicTents & I3D       & norm      & 0.004 & 0.012 & 25.0 & 0.20 & 0.65 & 0.04 & 0.21 \\
EpicTents & VideoMAE  & lr        & 0.008 & 0.032 & 19.4 & \textbf{0.51} & 0.64 & 0.06 & 0.24 \\
EpicTents & VideoMAE  & noise     & 0.004 & 0.200 & 16.0 & \textbf{0.51} & 0.65 & \textbf{0.14} & \textbf{0.34} \\
EpicTents & VideoMAE  & norm      & 0.004 & 0.047 & 16.0 & {0.49} & {\textbf{0.68}} & 0.12 & \textbf{0.34} \\

\midrule

IKEA & I3D       & lr        & 0.010 & 0.047 & 20.5 & 0.62 & 0.77 & 0.17 & 0.27 \\
IKEA & I3D       & noise     & 0.004 & 0.001 & 20.5 & 0.62 & 0.71 & 0.17 & 0.24 \\
IKEA & I3D       & norm      & 0.004 & 0.046 & 16.0 & 0.59 & 0.73 & 0.15 & 0.27 \\
IKEA & VideoMAE  & lr        & 0.005 & 0.047 & 20.5 & 0.58 & 0.75 & \textbf{0.28} & 0.34 \\
IKEA & VideoMAE  & noise     & 0.004 & 0.000 & 20.2 & \textbf{0.65} & {\textbf{0.79}} & 0.26 & 0.34 \\
IKEA & VideoMAE  & norm      & 0.004 & 0.046 & 16.0 & 0.63 & 0.78 & 0.24 & {\textbf{0.36}} \\
\midrule

EgoExo4D & I3D       & lr        & 0.0100 & 0.047 & 20.5 & 0.36 & 0.55 & 0.28 & 0.43 \\
EgoExo4D & I3D       & noise     & 0.0044 & 0.100 & 20.5 & 0.41 & 0.56 & 0.29 & 0.55 \\
EgoExo4D & I3D       & norm      & 0.0044 & 0.047 & 16.0 & 0.40 & 0.56 & 0.25 & 0.45 \\
EgoExo4D & VideoMAE  & lr        & 0.0005 & 0.047 & 20.5 & 0.44 & 0.58 & {\textbf{0.43}} & 0.50 \\
EgoExo4D & VideoMAE  & noise     & 0.0044 & 0.075 & 20.5 & 0.42 & \textbf{0.62} & 0.38 & 0.46 \\
EgoExo4D & VideoMAE  & norm      & 0.0044 & 0.047 & 25.0 & {\textbf{0.44}} & 0.64 & 0.33 & \textbf{0.53} \\

\bottomrule
\end{tabular}
\caption{
Grid search results for the USDL baseline. \\
\(\theta\): learning rate; \(\sigma\): augmentation noise std; \(\tau\): normalization parameter; \(\rho\): Spearman’s correlation. \\
Aggregation (Aggregated by): learning rate (lr), augmentation noise (noise), normalization (norm). \\
\textbf{Bold} highlights the best value per aggregation group within each dataset.
}
\label{tab:usdl_gridsearch_styled}
\end{table*}

\section{Qualitative Examples}

\begin{figure}[t!]
    \centering
    \begin{subfigure}[b]{0.9\linewidth}
        \includegraphics[width=\linewidth]{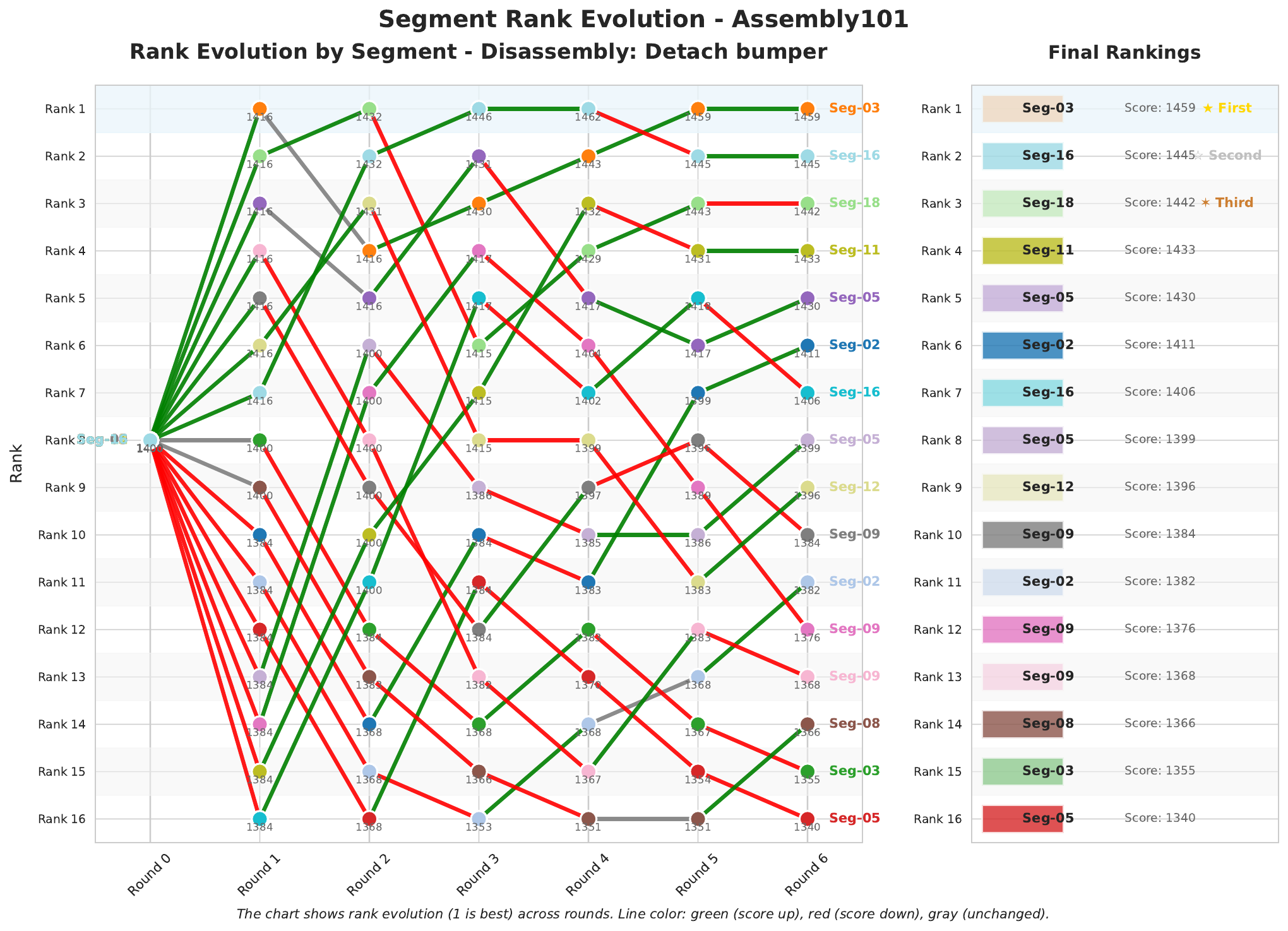}
        \caption{\textit{Assembly101} -- Detach Bumper}
    \end{subfigure}
    \vskip 0.5em
    \begin{subfigure}[b]{0.9\linewidth}
        \includegraphics[width=\linewidth]{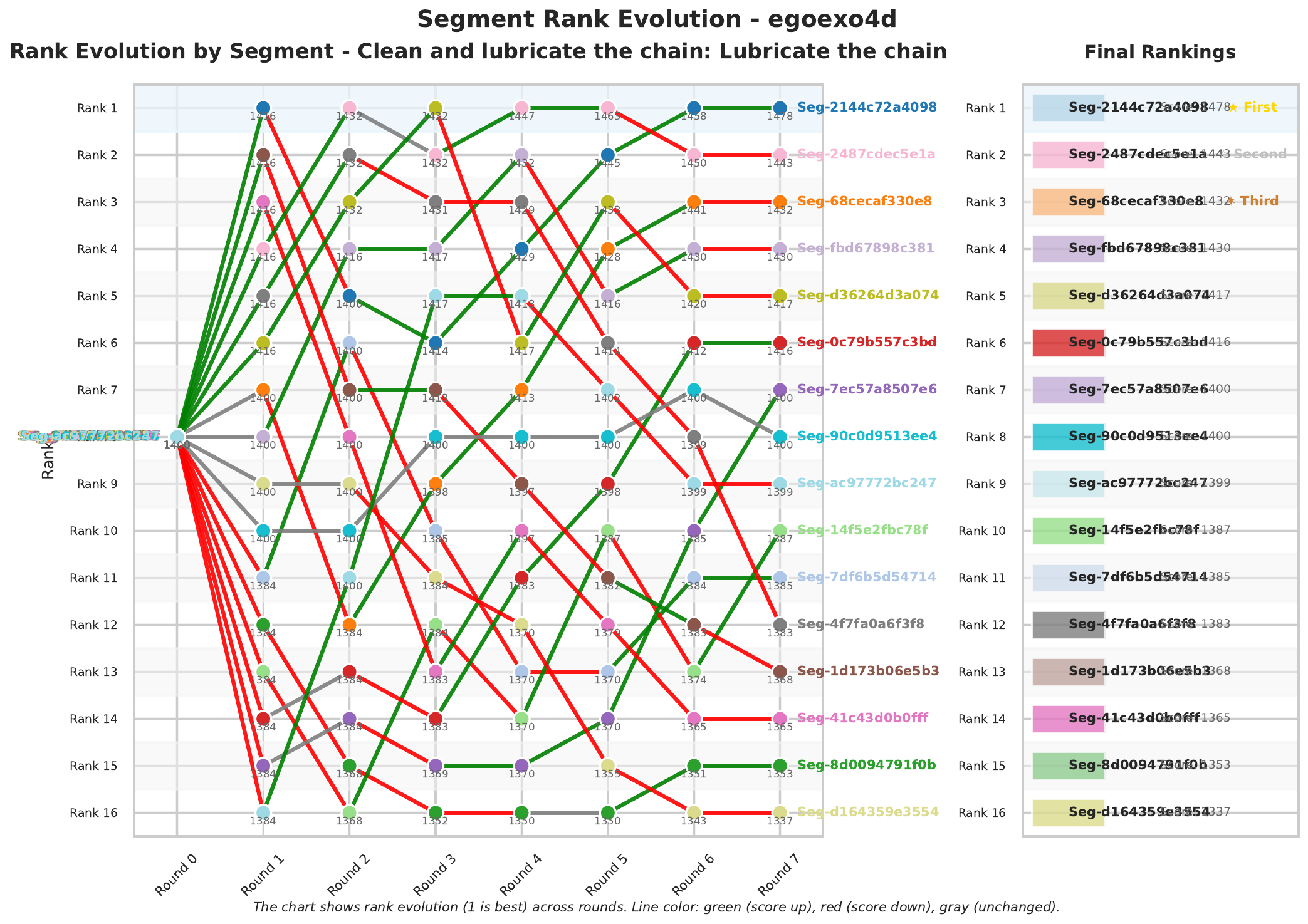}
        \caption{\textit{EgoExo4D} -- Lubricate the Chain}
    \end{subfigure}
    \caption{Ranking evolution of individual segments over Swiss tournament rounds. Each line corresponds to a segment, with the y-axis representing its rank (1 is best). Color indicates whether the segment's final score increased (green), decreased (red), or remained unchanged (gray) compared to earlier rounds.}
    \label{fig:qualitative_rank_evolution}
\end{figure}

\noindent To better understand how segment rankings evolve across Swiss tournament rounds, we provide qualitative examples from two datasets in Figure~\ref{fig:qualitative_rank_evolution}. Each plot illustrates the rank trajectory of all segments involved in a specific sub-task.
\noindent In the \textit{Assembly101} example, we observe clear separation among high-performing segments early in the process, with minimal rank fluctuation in the final rounds. In contrast, \textit{EgoExo4D} exhibits a high degree of stability for both top- and bottom-ranked segments from as early as round 3, confirming the Swiss format's effectiveness in identifying the extremes of the ranking spectrum with limited supervision.

\noindent These examples visually reinforce the earlier quantitative findings: although only a subset of all possible comparisons is used, the Swiss tournament efficiently converges to rankings that closely approximate the full round-robin outcome.

\end{appendices}

\end{document}